\documentclass[authoryear]{elsarticle}

\usepackage{hyperref}

\usepackage[T1]{fontenc}
\usepackage[latin9]{inputenc}
\usepackage{float}
\usepackage{amsmath}
\usepackage{amssymb}
\usepackage{amsthm}
\usepackage{graphicx}

\usepackage{color,soul}
\usepackage[author={FMZennaro}]{pdfcomment}

\newcommand{\revvv}[2]{#2}

\makeatletter
\def\ps@pprintTitle{%
	\let\@oddhead\@empty
	\let\@evenhead\@empty
	\def\@oddfoot{}%
	\let\@evenfoot\@oddfoot}
\makeatother

\makeatletter
\newenvironment{lyxlist}[1]
{\begin{list}{}
{\settowidth{\labelwidth}{#1}
 \setlength{\leftmargin}{\labelwidth}
 \addtolength{\leftmargin}{\labelsep}
 }}
{\end{list}}
\makeatother

\DeclareMathOperator*{\argmax}{\arg\!\max}

\newtheorem{theorem}{Theorem}
\newtheorem{proposition}{Proposition}
\newtheorem{lemma}{Lemma}

\biboptions{authoryear}

\begin{document}

\begin{frontmatter}

\title{Towards Understanding Sparse Filtering: A Theoretical Perspective}

\author[mymainaddress]{Fabio Massimo Zennaro\corref{mycorrespondingauthor}}
\cortext[mycorrespondingauthor]{Corresponding author}
\ead{zennarof@cs.manchester.ac.uk}

\author[mymainaddress]{Ke Chen}
\ead{chen@cs.manchester.ac.uk}

\address[mymainaddress]{School of Computer Science, The University of Manchester, Manchester, M13 9PL, UK}

\begin{abstract}
In this paper we present a theoretical analysis to understand sparse filtering, a recent and effective algorithm for unsupervised learning. 
The aim of this research is not to show \emph{whether} or \emph{how well} sparse filtering works, but to understand \emph{why} and \emph{when} sparse filtering does work. 
We provide a thorough theoretical analysis of sparse filtering and its properties, and further offer an experimental validation of the main outcomes of our theoretical analysis. 
We show that sparse filtering works by explicitly maximizing the entropy of the learned representations through the maximization of the proxy of sparsity, and by implicitly preserving mutual information between original and learned representations through the constraint of preserving a structure of the data. Specifically, we show that the sparse filtering algorithm implemented using an absolute-value non-linearity determines the preservation of a data structure defined by \revvv{I rephrased this to refer to cosine distance (which is a known concept) instead of cosine neighborhoodness (which is no equivalently well known)}{ relations of neighborhoodness under the \emph{cosine distance}}. 
Furthermore, we empirically validate our theoretical results with artificial and real data sets, and we apply our theoretical understanding to explain the success of sparse filtering on real-world problems. 
Our work provides a strong theoretical basis for understanding sparse filtering: it highlights assumptions and conditions for success behind this feature distribution learning algorithm, and provides insights for developing new feature distribution learning algorithms.
\end{abstract}


\end{frontmatter}


\section{Introduction}

Unsupervised learning deals with the \emph{problem of modeling data}, stated as the problem of learning a transformation which maps data in a given representation onto a new representation.
Contrasted with supervised learning, where we are provided labels and we learn a relationship between the data and the labels, unsupervised learning does not rely on any provided external semantics in the form of labels. In order to learn, unsupervised learning relies on the specification of assumptions and constraints that express our very understanding of the problem of modeling the data; for example, if we judge that a useful representation of the data would be provided by grouping together data instances according to a specific metric, then we may rely on distance-based clustering algorithms to generate one-hot representations of the data. \\

Often, the tacit aim of unsupervised learning is to generate representations of the data that may simplify the further problem of learning meaningful relationships through supervised learning. \citet{Coates2011} clearly showed that very simple unsupervised learning algorithms (such as $k$-means clustering), when properly tuned, can generate representations of the data that allow even basic classifiers, such as a linear support vector machine, to achieve state-of-the-art performances.\\

One common assumption hard-wired in several unsupervised learning
algorithms is \emph{sparsity} \citep[for a review on the use of sparsity in representation learning see ][]{Bengio2012}. Sparse representation learning aims at finding a mapping that produces new representations where few of the components are active while all of the others are reduced to zero. The adoption of sparsity relies both on biological analogies and on theoretical justifications \citep[for discussion on the justification of sparsity see, for instance, ][]{ Foldiak1995, Olshausen1997, Ganguli2012, Bengio2012}. Several state-of-the-art algorithms have been developed or have been adapted to learn sparse representations  \citep[for a recent survey of these algorithms, see ][]{Zhang2015a}.

\subsection{Sparse Filtering and Related Work \label{sub:State-art}}

In 2011, \citet{Ngiam2011} proposed a novel unsupervised learning framework for generating sparse representations. 
Most of the successful unsupervised algorithms may be described as \emph{data distribution learning} algorithms  that try to learn new representations which better model the underlying probability distribution that generated the data. In contrast, they proposed the possibility of developing \emph{feature distribution learning} algorithms  that try to learn a new representations having desirable properties, without the need of taking into account the problem of modeling the distribution of the data.

Consistently with the feature distribution learning framework, they defined an algorithm named \emph{sparse filtering}, which ignores the problem of learning the data distribution and instead focuses only on optimizing the sparsity of the learned representations. 
Sparse filtering proved to be an excellent algorithm for unsupervised learning: it is extremely simple to tune since it has only a single hyper-parameter to select; it scales very well with the dimension of the input; it is easy to implement; and, more importantly, it was shown to achieve state-of-the-art performance on image recognition and phone classification \citep{Ngiam2011, Goodfellow2013, Romaszko2013}.
Thanks to its success and to the simplicity of implementing and integrating the algorithm in already existing  machine learning systems, sparse filtering was adopted in many real-world applications  \citep[see, for instance, the works of ][]{Dong2014, Raja2015, Lei2015, Ryman2016}.

Some studies have also provided sparse filtering with some biological
support. 
\citet{Bruce2015} analyzed different biologically-grounded principles for representation learning of images, using sparse filtering as a starting point for the definition of new learning algorithms.
Interestingly, \citet{Kozlov2016} used sparse filtering to model the receptive fields of high-level auditory neurons in the European starling, providing further support to the general hypothesis that sparsity and normalization are general principles of neural computation \citep{Carandini2012}.

\subsection{Problem Statement}

So far, sparse filtering has been successfully applied to many scenarios, and its usefulness repeatedly confirmed \citep[see, for instance, its application in ][]{Dong2014, Raja2015, Han2016a, Liu2016}. 
In general, however, a clear theoretical explanation of the algorithm is still lacking. \citet{Ngiam2011} drew connections between sparse filtering, divisive normalization, independent component analysis, and sparse coding, while \citet{Lederer2014} provided a deeper analysis of the normalization steps inside the sparse filtering algorithm.
However, the reasons why and on what conditions sparse filtering works are left unexplored. 
In this paper, we aim at understanding from a theoretical perspective \emph{why} and \emph{when} sparse filtering works. It is worth clarifying that our work does not concern itself with showing \emph{whether} or \emph{how well} well sparse filtering works, as there have been abundant evidence in literature on its successes in different real applications. \\

We begin by arguing that any unsupervised learning algorithm, in order to work properly, has to deal with the problem of preserving information conveyed by the probability distribution of the data. Given that feature distribution learning ignores the problem of learning the data distribution itself, a natural question arises: \emph{how is the information conveyed by the data distribution preserved in feature distribution learning and, specifically, in sparse filtering?}\\

The actual success of sparse filtering suggests that the algorithm is indeed able to preserve relevant information conveyed in the distribution of the data. However, no explanation for this behavior has been given. We suggest that information may be preserved through the preservation of the structure of the data.  To understand how this may be, we study the properties of the transformations within the algorithm and pose the following question: \emph{is there any sort of data structure that is preserved by the processing in sparse filtering?} \\

Through a theoretical analysis we show that sparse filtering implemented using an absolute-value non-linearity does indeed retain information through the preservation of the data structure defined by the relations of neighborhoodness under the cosine distance. Relying on this, we investigate whether our theoretical results can be used to explain the success or the failure of sparse filtering in real applications. In particular we consider the following questions: \revvv{Following Rev3 I rephrased this using an impersonal form}{}\emph{can the success of sparse filtering be explained in terms of the type of structure preserved? Can the failure of alternative forms of sparse filtering using different non-linearities be explained counterfactually on the grounds of information preservation? Is it possible to identify scenarios in which sparse filtering is likely to be helpful and other scenarios in which it is likely not to be useful? }

\subsection{Contributions}

We summarize the contributions made in this study as follows:
\begin{itemize}
\item We provide a theoretical analysis to understand why and when sparse filtering works. 
We show that the standard sparse filtering algorithm implemented with an absolute-value non-linearity implicitly works under the assumption of an intrinsic radial structure of the data. This assumption naturally makes the algorithm more suitable for certain data sets.
\item We empirically validate our main theoretical findings, both on artificial data and real-world data sets.
\item We provide useful insights for developing new feature distribution learning algorithms based on our theoretical understanding.
\end{itemize}

\subsection{Organization}

The rest of this paper is organized as follows. 
We first review the concepts and ideas forming the foundations of our work (Section \ref{sec:Foundations}).
Next, we provide a formal theoretical analysis of the sparse filtering algorithm based on a rigorous conceptualization of feature distribution learning (Section \ref{sec:Theoretical-Analysis-of}).
The theoretical results inform the following experimental simulations (Section \ref{sec:Experimental-Simulation}). 
We then discuss the results we collected, in relation to sparse filtering, in particular, and to feature distribution learning, in general (Section \ref{sec:Discussion}). 
Finally, we draw conclusions by summarizing our contributions and highlighting future developments (Section \ref{sec:Conclusion}).

To facilitate our presentation, Table \ref{tab:Notation.} summarizes the notation system used in this manuscript.

\begin{table}
\begin{centering}
\rule[0.5ex]{1\columnwidth}{1pt}
\begin{lyxlist}{00.00.0000}
	\begin{small}
\item [{$N$}] Number of samples.
\item [{$O$}] Original dimensionality of the samples.
\item [{$L$}] Learned dimensionality of the samples.

\item [{$\mathbf{X}$}] Matrix of original representations with domain
$\mathbb{R}^{O\times N}.$
\item [{$\mathsf{X}$}] Data set or collection of data.
\item [{$\mathbf{X}^{(i)}$}] $i$-th sample from $\mathbf{X}$; vector
of shape $(O\times1)$ with domain $\mathbb{R}^{O}$, $1\leq i\leq N$. 
\item [{$\mathbf{X}_{j}$}] $j$-th feature from $\mathbf{X}$; vector
of shape $(1\times N)$ with domain $\mathbb{R}^{N}$, $1\leq j\leq O$. 
\item [{$\mathbf{X}_{j}^{(i)}$}] $j$-th feature of the $i$-th sample
from $\mathbf{X}$; scalar with domain $\mathbb{R}$. 
\item [{$X$}] Multivariate random variable (random vector) modeling the
original data $\mathbf{X}^{(i)}.$
\item [{$p(X)$}] Probability density function of the original representations.
\item [{$p\left(\mathbf{X}^{(i)}\right)$}] Probability of the outcome
$\mathbf{X}^{(i)}$ when sampling from $p(X)$.\\
$p\left(\mathbf{X}^{(i)}\right)$ is the shorthand for the more rigorous
notation $p\left(X=\mathbf{X}^{(i)}\right)$.

\item [{$\mathbf{Z}$}] Matrix of learned representations with domain $\mathbb{R}^{L\times N}.$ 
\item [{$\mathbf{F}/\mathbf{\tilde{F}}/\mathbf{\hat{F}}$}] Matrix of intermediate
representations with domain $\mathbb{R}^{L\times N}.$ 
\item [{$\mathbf{Y}$}] Vector of labels associated with the data with
domain $\mathbb{R}^{1\times N}.$
\item [{$\mathbf{W}$}] Matrix of weights with domain domain $\mathbb{R}^{L\times O}.$ 

\end{small}
\end{lyxlist}
\rule[0.5ex]{1\columnwidth}{1pt}
\end{centering}

\caption{Notation. \label{tab:Notation.}}
\end{table}

\section{Foundations\label{sec:Foundations}}

In this section we review basic concepts underlying our study. We provide a rigorous description of unsupervised learning, we present its formalization in information-theoretic terms, we formalize the property of sparsity, and, finally, we bring all these concepts together in the definition of the sparse filtering algorithm.

\subsection{Unsupervised Learning \label{sec:unsupervised_learning}}

Let $\mathsf{X}=\{\mathbf{X}^{(i)}\in\mathbb{R}^{O}\}_{i=1}^{N}$
be a set of $N$ \emph{samples} or \emph{data points} represented
as vectors in an $O$-dimensional space. We will refer to the given
representation of a sample $\mathbf{X}^{(i)}$ in the space $\mathbb{R}^{O}$
as the \emph{original representation} of the sample $\mathbf{X}^{(i)}$
and to $\mathbb{R}^{O}$ as the \emph{original space}. From an algebraic point of view, we can formalize the data set as
a matrix $\mathbf{X}$ of dimensions $(O\times N)$; from a probabilistic
point of view, we can model the data points $\mathbf{X}^{(i)}$ as
i.i.d. samples from a multivariate random variable $X=\left(X_{1},X_{2},\ldots,X_{O}\right)$
with pdf $p\left(X\right)$.

Unsupervised learning discovers a transformation $f:\,\mathbb{R}^{O}\rightarrow\mathbb{R}^{L}$
mapping the set $\mathsf{X}$
from an $O$-dimensional space to the set $\mathsf{Z}=\{\mathbf{Z}^{(i)}\in\mathbb{R}^{L}\}_{i=1}^{N}$
in an $L$-dimensional space. We will refer to the transformed representation
$\mathbf{Z}^{(i)}$ in the space $\mathbb{R}^{L}$ as the \emph{learned
representation} of the sample $\mathbf{X}^{(i)}$ and to $\mathbb{R}^{L}$
as the \emph{learned space}. Again, from an algebraic point of view,
we can formalize the transformed data set as a matrix $\mathbf{Z}$
of dimensions $(L\times N)$; from a probabilistic point of view, we
can model the data points $\mathbf{Z}^{(i)}$ as i.i.d. samples from
a multivariate random variable $Z=\left(Z_{1},Z_{2},\ldots,Z_{L}\right)$
with pdf $p\left(Z\right)$.

Unsupervised learning is often used for learning better representations
for ensuing supervised tasks. Suppose that we are given a set $\mathsf{Y}=\{\mathbf{Y}^{(i)}\in\mathbb{R}\}_{i=1}^{N}$
of $N$ labels, such that the $i^{th}$ label in $\mathsf{Y}$
is associated to the $i^{th}$ sample in $\mathsf{X}$. From an algebraic
point of view, we can formalize the labels as a vector $\mathbf{Y}$
of dimensions $(1\times N)$; from a probabilistic point of view, we
can model the labels $\mathbf{Y}^{(i)}$ as i.i.d. samples from a random variable
$Y$ with pdf $p\left(Y\right)$. Let us now consider the new data
set $\left(\mathsf{X},\mathsf{Y}\right)=\left\{ \left(\mathbf{X}^{(i)},\mathbf{Y}^{(i)}\right)\in\mathbb{R}^{O}\times\mathbb{R}\right\} _{i=1}^{N}$.
In this scenario, the aim of unsupervised learning is to learn from $\mathbf{X}^{(i)}$ representations
$\mathbf{Z}^{(i)}$ such that modeling the relationship $g':\,\mathbf{Z}^{(i)}\mapsto \mathbf{Y}^{(i)}$ or the distribution $P(Y\vert Z)$
is easier than modeling the relationship $g'':\,\mathbf{X}^{(i)}\mapsto \mathbf{Y}^{(i)}$ or the distribution $P(Y \vert X)$.\\

\textbf{Clustering.} A specific form of unsupervised learning is \emph{clustering}.
  
\emph{Hard clustering} discovers a transformation
$f:\,\mathbb{R}^{O}\rightarrow\mathbb{R}^{L}$ mapping the original
samples $\mathbf{X}^{(i)}$ onto one-hot representations $\mathbf{Z}^{(i)}$,
where the single non-null component of $\mathbf{Z}^{(i)}$ encodes the
assignment of the original sample to a cluster.

\emph{Soft clustering} discovers
a transformation $f:\,\mathbb{R}^{O}\rightarrow\mathbb{R}^{L}$ mapping
the original samples $\mathbf{X}^{(i)}$ onto representations $\mathbf{Z}^{(i)}$,
where the value of each component of $\mathbf{Z}^{(i)}$ encodes the
degree of membership of the original sample to each cluster. Soft clustering algorithms may be used for learning representations
$\mathbf{Z}^{(i)}$ that simplify the problem of modeling the relationship
$g':\,\mathbf{Z}^{(i)}\mapsto \mathbf{Y}^{(i)}$; in this case, the soft clustering
algorithm is normally grounded in the following assumptions. 
(i) Samples
are taken to be first generated by a stochastic process with pdf $p\left(X^{*}\right)$; the samples are corrupted by various forms of noise; the noisy samples that we receive as original representations $\mathbf{X}^{(i)}$ follow a noisy pdf $p(X)$; the noiseless distribution
underlying the data is referred to as \emph{true} pdf $p\left(X^{*}\right)$. 
(ii) Noiseless samples generated by the true pdf $p\left(X^{*}\right)$ are
taken to have a stronger correlation to the labels $\mathbf{Y}^{(i)}$ than 
the original samples $\mathbf{X}^{(i)}$. 
(iii) The true pdf $p\left(X^{*}\right)$ may be approximated through a mixture model.
(iv) Relationships of neighborhoodness under a chosen metric in the
original space $\mathbb{R}^{O}$ allows us to recover the true
pdf $p\left(X^{*}\right)$. 
Based on these assumptions, soft clustering
algorithms instantiate a set of $C$ clusters (each one describing
one component of the mixture model) and group into clusters nearby data
points. Two data points $\mathbf{X}^{(1)}$ and $\mathbf{X}^{(2)}$
falling in the same clusters are represented by the same exemplar
$\bar{\mathbf{X}}$, assuming that such an exemplar contains all the
relevant information carried by $\mathbf{X}^{(1)}$ and $\mathbf{X}^{(2)}$,
and that the information contained in the difference between $\mathbf{X}^{(1)}$
or $\mathbf{X}^{(2)}$ and the exemplar $\bar{\mathbf{X}}$ amounts
to noise. 
If the assumptions are correct, a soft clustering algorithm
will learn new representations $\mathbf{Z}^{(i)}$ whose pdf $p(Z)$
is closer to the true pdf $p\left(X^{*}\right)$ than the original
pdf $p(X)$; therefore, it will be easier to learn $g':\,\mathbf{Z}^{(i)}\mapsto Y^{(i)}$
or $p(Y\vert Z)$ than learning $g'':\,\mathbf{X}^{(i)}\mapsto \textbf{Y}^{(i)}$
or $p(Y\vert X)$.\\

\textbf{Distribution Learning.} Another form of unsupervised learning is \emph{distribution learning}.

\emph{Data distribution learning} is a generic term for algorithms that aim at estimating the true pdf $p\left(X^{*}\right)$ from the available data. Examples of data distribution learning algorithms include \citep{Ngiam2011}: denoising auto-encoders (DAE) \citep{Vincent2008}, restricted Boltzmann machines (RBM) \citep{Hinton2006b}, and independent component analysis (ICA) \citep{Bell1997}. 
 In the context of learning for supervised tasks, 
if we learn a pdf $p(Z)$ that well approximates the true pdf $p\left(X^{*}\right)$, we can reasonably expect that the ensuing learning of $p(Y\vert Z)$ will be simplified \citep{Bengio2012}.

\emph{Feature distribution learning}, in contrast, denotes algorithms aimed at learning
a pdf $p\left(Z\right)$ which has a set of desirable properties. It overlooks the problem of estimating the true distribution $p\left(X^{*}\right)$ and focuses instead on shaping the learned pdf $p\left(Z\right)$ according to chosen criteria.
The most representative algorithm of this family is sparse filtering
(SF) \citep{Ngiam2011}. In the context of learning for supervised tasks, learning a pdf $p(Z)$ with specific properties is meaningful if we know a priori that certain properties (such as sparsity or smoothness) will be useful for supervised learning.

\subsection{Information-Theoretic Aspects of Unsupervised Learning \label{ITL}}

Relying on conceptual tools from information theory, \citet{Vincent2010} argued that an unsupervised learning algorithm can generate good representations by satisfying two requirements: (i) retaining information about the input, and (ii) applying constraints that lead to the extraction of useful information from noise.

In more general terms, we may state that a good unsupervised representation may be obtained by satisfying the two following information-theoretic requirements:  (i) maximizing the mutual information between input and output \citep[\emph{infomax principle}, ][]{Linsker1989}, and (ii) maximizing a measure of information of the output (\emph{informativeness principle}\footnote{We named this principle \emph{informativeness principle} for lack of a better term.}). 

As such, in order to generate good representations, an unsupervised learning algorithm has to somehow negotiate the trade-off between the infomax principle and the informativeness principle:
\begin{equation}
\underbrace{\argmax_{p(Z) \in \mathcal{P}}\,D\left[p(X,Z)\parallel p(X)p(Z)\right]}_{\begin{array}{c}
	\small\textnormal{infomax principle}	
	\end{array}}+\underbrace{\argmax_{p(Z) \in \mathcal{P}}\,D\left[p(Z)\parallel q\right]}_{\begin{array}{c}
	\small \textnormal{informativeness principle}
	\end{array}}.\label{eq:ITL formula}
\end{equation}
where $D\left[\cdot\right]$ is a measure of distance or divergence
between pdfs, such as the Kullback-Leibler divergence \citep{MacKay2003}, $q$ is an entropy-maximizing pdf, and $\mathcal{P}$ is the space of all the pdfs defined on the space of the learned representations $\mathbf{Z}$.

Maximizing the infomax principle may be expressed as the maximization
of the mutual information $I\left[X;Z\right]$, or, equivalently, as
the maximization of the relative entropy between $p(X,Z)$ and $p(X)p(Z)$.
Maximizing the informativeness may be expressed as the minimization
of the entropy $H\left[Z\right]$ or the maximization of the relative
entropy between the learned pdf $p\left(Z\right)$ and the entropy-maximizing
pdf $q$.

Unfortunately, the objective defined in Equation \ref{eq:ITL formula} is bound to remain mainly theoretical, as information-theoretic quantities are extremely hard to estimate in practice. Therefore we need to rely on approximations or heuristics to make these quantities tractable.

\subsection{Sparsity}

Given a generic vector $\mathbf{v}$ in an $N$-dimensional space,  $\mathbf{v}$ is sparse if a small number of components of the vector accounts for most of the energy of the vector \citep{Hurley2009}.
Practically,  the vector $\mathbf{v}$ is \emph{sparse} if $n\ll N$ components of the vector $\mathbf{v}$ are active (that is, have a value different from zero) while the remaining $N-n$ components are inactive (that is, have the value zero). A vector $\mathbf{v}$ is \emph{$k$-sparse} if exactly $k$ components are active. By analogy, we may define sparsity for matrices (with reference to their components) and for random variables (with reference to their realizations). 

Several measures of sparsity have been proposed in the literature \citep[for a review of different measures of sparsity and their properties, see ][]{Hurley2009}.
 A common family of measures of sparsity is the $\ell_{p}$-norm family: $\ell_{p}(\mathbf{v})= \sqrt[p]{\sum_{i=1}^{N}\left|\mathbf{v}_{i}\right|^p}$. 
The most intuitive measure is the $\ell_0$-norm which computes the number of non-zero components of a vector; however, this measure is practically inadequate, as in concrete implementations the components of a vector are rarely reduced perfectly to zero. The simplest relaxation of the $\ell_0$-norm is the $\ell_1$-norm, which is often referred to as \emph{activation} in the sparse filtering literature. \revvv{I clarify here that it is the negative form of l1 to be the exact proxy of l0.}{} The negative form of the $\ell_1$-norm works as an efficient proxy for measuring the $\ell_0$-norm \citep{Elad2010}. Given a representation $\mathbf{Z}^{(i)}$, $\ell_{1}\left(\mathbf{Z}^{(i)}\right)$ or $activation\left(\mathbf{Z}^{(i)}\right)$ quantifies the sparsity of $\mathbf{Z}^{(i)}$. Minimizing the activation of the learned representation $\mathbf{Z}^{(i)}$ will maximize the $\ell_0$-norm and the sparsity of $\mathbf{Z}^{(i)}$.

\subsection{Sparse Filtering }

Sparse filtering is the most representative example of feature distribution learning algorithms \citep{Ngiam2011}. Its aim is learning a pdf $p(Z)$ which maximizes the sparsity of the learned representations $\mathbf{Z}^{(i)}$.\\

\textbf{Enforcement of sparsity in sparse filtering.} Sparse learned representations $\mathbf{Z}^{(i)}$ are achieved by enforcing three constraints on the matrix of learned representations
$\mathbf{Z}$: 
\begin{itemize}
\item \emph{Population sparsity}: each sample $\mathbf{Z}^{(i)}$, 
is required to be sparse, that is, described only by a few features.
The sparsity of a sample $\mathbf{Z}^{(i)}$ is computed as its activation:
$\ell_{1}\left(\mathbf{Z}^{(i)}\right)=\sum_{j=1}^{L}\left|\mathbf{Z}_{j}^{(i)}\right|$.
\item \emph{Lifetime sparsity}: each feature $\mathbf{Z}_{j}$, 
is required to be sparse, that is, to describe only a few samples. Lifetime
sparsity is often referred to as \emph{selectivity} \citep{Goh2012}.
The sparsity of a feature $\mathbf{Z}_{j}$ is computed as its activation:
$\ell_{1}\left(\mathbf{Z}_{j}\right)=\sum_{i=1}^{N}\left|\mathbf{Z}_{j}^{(i)}\right|$.
\item \emph{High dispersal}: all the features are required to have approximately 
the same activation. The dispersal of the features is computed as the variance of the
activation across all the features: $Var\left[activation\left(\mathbf{Z}_{j}\right)\right]=E\left[\ell_1\left(\mathbf{Z}_{j}\right)^{2}\right]-E\left[\ell_1\left(\mathbf{Z}_{j}\right)\right]^{2}$. Lower variance
corresponds to higher dispersal.
\end{itemize}
The enforcement of these three properties translates into learning
non-degenerate sparse representation. \\

\begin{figure}
\begin{centering}
\includegraphics[scale=0.48]{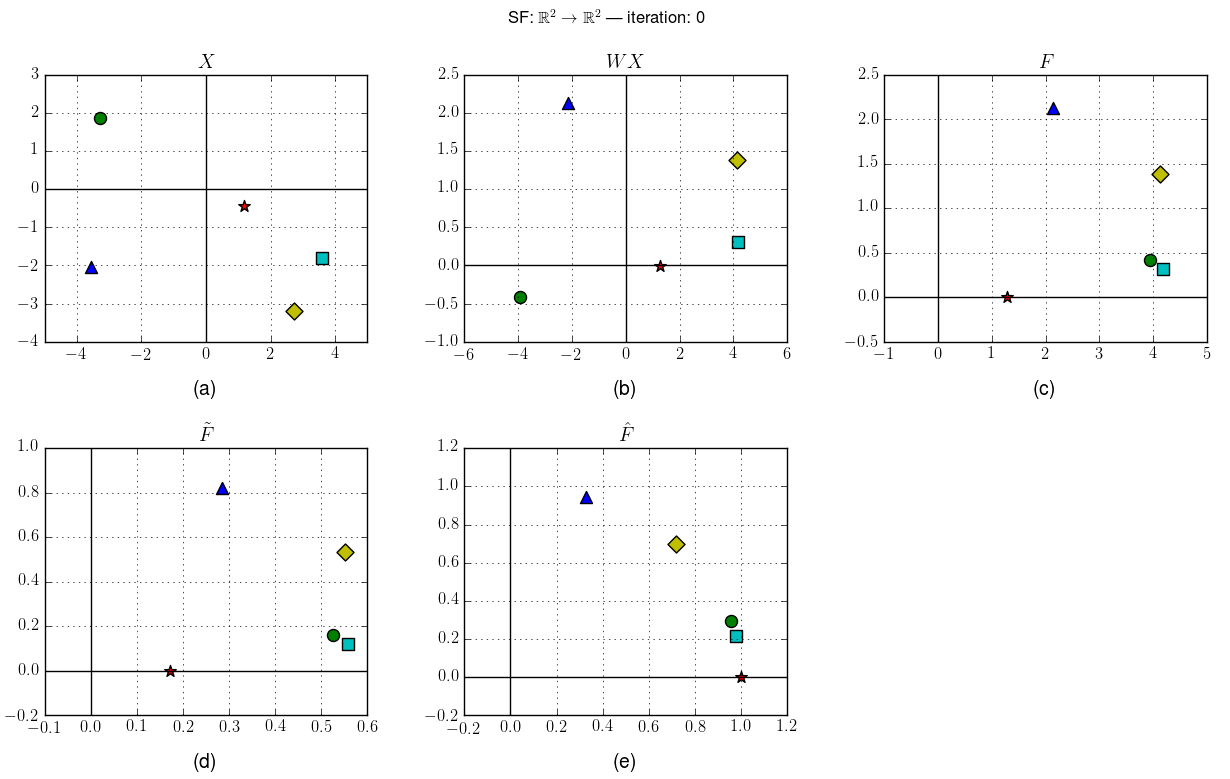}
\par\end{centering}

\caption{Illustration of sparse filtering.\protect \\
Sparse filtering is applied to a random set of data $\mathsf{X}$ constituted by the matrix $\mathbf{X}$ containing
five samples ($N=5$) in two dimensions ($O=2$). Each point is
generated by sampling its coordinates from a uniform distribution
$Unif\left(-5,5\right)$. Sparse filtering is used to learn a new
representation of the data in two dimensions ($L=2$). This figure
shows the transformations determined by the sparse filtering algorithm
at iteration $0$, after the weight matrix $\mathbf{W}$ has been
randomly initialized and before any training.\protect \\
(a) Original representation of the data $\mathbf{X}$ in $\mathbb{R}^{2}$.
(b) Linear projection of the data onto the intermediate representation
$\mathbf{WX}$. (c) Non-linear projection of $\mathbf{WX}$ using
an absolute-value function onto the intermediate representation $\mathbf{F}$.
(d) $\ell_{2}$-normalization of the data $\mathbf{F}$ along the
features, yielding the intermediate representation $\tilde{\mathbf{F}}$.
(e) $\ell_{2}$-normalization of the data $\tilde{\mathbf{F}}$ along
the samples, yielding the final learned representation $\mathbf{\hat{F}}=\mathbf{Z}$.\protect \\
Notice that the colors and the markers of the data points $\mathbf{X}^{(i)}$ do not have any meanings. A random color and marker have been assigned to each point in order to allow the tracking of the location of the points through the different transformations applied by sparse filtering. \label{fig:Illustration-of-SF}}
\end{figure}

\textbf{Sparse filtering algorithm.} Sparse filtering is implemented as a simple algorithm in six steps
(refer to Figure \ref{fig:Illustration-of-SF}
for an illustration of the transformations on a two-dimensional data set): 

\renewcommand{\labelenumi}{A\arabic{enumi}.}

\begin{enumerate}
\setcounter{enumi}{-1}
\item Initialization of the weights: the weight matrix $\mathbf{W}$ is initialized  sampling each component from a normal distribution $\mathcal{N}(0,1)$. 
\end{enumerate}
\begin{enumerate}
\item Linear projection of the original data: $f_{A1}(\mathbf{X})=\mathbf{W}\mathbf{X}$.
The weight matrix $\mathbf{W}$ can be interpreted as a \emph{dictionary}
\citep{Denil2012} or as a\emph{ filter bank} \citep{Dong2015}, where
each row is a codeword or a filter applied to each sample. Refer to Figure \ref{fig:Illustration-of-SF}(a)
and \ref{fig:Illustration-of-SF}(b) for an illustration of this transformation.
\item Non-linear transformation: $\mathbf{F}=f_{A2}\left(\mathbf{W}\mathbf{X}\right)$,
where $f_{A2}(\cdot):\mathbb{R}\rightarrow\mathbb{R}$ is an element-wise
non-linear function. Although this non-linear function can, in principle,
be arbitrarily chosen, all the implementations known to the authors
used an element-wise absolute-value function $f(x)=\left|x\right|$.
For practical reasons, this non-linearity is implemented as a soft
absolute-value $f(x)=\sqrt{x^{2}+\epsilon}$, where $\epsilon$ is
a small negligible value (for instance, $\epsilon=10^{-8}$). Refer to Figure
\ref{fig:Illustration-of-SF}(b) and \ref{fig:Illustration-of-SF}(c)
for an illustration of this transformation.
\item $\ell_{2}$-normalization along the features (or along the rows):
$\tilde{\mathbf{F}}=f_{A3}\left(\mathbf{F}\right)=\left[ \frac{\mathbf{F}_{j}^{(i)}}{\sqrt{\sum_{i=1}^{N}\left(\mathbf{F}_{j}^{(i)}\right)^{2}}}\right] $.
In this step, each feature is normalized so that its squared activation
is one, that is, $\sum_{i=1}^{N}\left(\tilde{\mathbf{F}}_{j}^{(i)}\right)^{2}=1$.
Refer to Figure \ref{fig:Illustration-of-SF}(c) and \ref{fig:Illustration-of-SF}(d)
for an illustration of this transformation. 
\item $\ell_{2}$-normalization along the samples (or along the columns):
$\mathbf{Z}=\hat{\mathbf{F}}=f_{A4}\left(\tilde{\mathbf{F}}\right)=\left[ \frac{\tilde{\mathbf{F}}_{j}^{(i)}}{\sqrt{\sum_{j=1}^{L}\left(\tilde{\mathbf{F}}_{j}^{(i)}\right)^{2}}}\right] $.
In this step, each sample is normalized so that its squared activation
is one, that is, $\sum_{j=1}^{L}\left(\hat{\mathbf{F}}_{j}^{(i)}\right)^{2}=1$.
Refer to Figure \ref{fig:Illustration-of-SF}(d) and \ref{fig:Illustration-of-SF}(e)
for an illustration of this transformation.
\item $\ell_{1}$-minimization: $ \displaystyle \min_{\hat{\mathbf{F}}\in\mathbb{R}^{L \times N}}$ $\sum_{ij}\hat{\mathbf{F}}_{j}^{(i)}$.
This minimization is the objective of sparse filtering; by minimizing
the overall activation of the matrix $\hat{\mathbf{F}}$, we maximize
the sparsity of the learned representations. 
\end{enumerate}
\renewcommand{\labelenumi}{\arabic{enumi}.}

After learning, new data $\mathbf{X}'$ is processed through step A1 to
A4, such that $\mathbf{Z}'=f_{A1:A4}\left(\mathbf{X}'\right)=f_{A4}\left(f_{A3}\left(f_{A2}\left(\mathbf{WX}'\right)\right)\right)$.

%

As explained by \citet{Ngiam2011}, the combination of the $\ell_{1}$-minimization
with the two $\ell_{2}$-normalizations guarantees the learning of
a representation with the properties of population sparsity, lifetime
sparsity, and high dispersal.

\section{Theoretical Analysis of Sparse Filtering\label{sec:Theoretical-Analysis-of}}

In order to better understand sparse filtering we first re-formulate and explain this algorithm in terms of information-theoretic concepts. Relying on this improved understanding, we will then move on to a formal analysis of the sparse filtering algorithm.

In detail, Section \ref{sec:info-theory} presents our thesis stating that sparse filtering must satisfy the informativeness and the infomax principle. Section \ref{sec:informativeness} shows how sparse filtering satisfies the informativeness principle, while Section \ref{sec:infomax} introduces the hypothesis that sparse filtering satisfies the infomax principle through the preservation of structure of the data. Section \ref{sec:no-euclid} rules out the simplest hypothesis that sparse filtering preserves a structure explained by the Euclidean metric. The following three sections show important properties of sparse filtering related to structure preservation: Section \ref{sec:pres-coll} proves that sparse filtering preserve collinearity, Section \ref{sec:homorepresentation-collinear} proves that collinear points are mapped onto identical representation, and finally, Section \ref{sec:homorepresentation-moduli} proves that points having the same moduli are mapped onto identical representations. Section \ref{sec:pres-cosine-neigh} puts together these results to conclude that sparse filtering preserves relations of cosine neighborhoodness 
Section \ref{sec:basis-pursuit} and Section \ref{sec:repr-filters} delve deeper in the dynamics of sparse filtering providing a geometric interpretation of the algorithm in terms of bases of the learned space and filters in the original space. 
Section \ref{sec:alternative-non-linearities} and Section \ref{sec:pres-euclidean} investigate the limits of the sparse filtering algorithm, by evaluating more closely the role of the absolute-value non-linearity in the preservation of structure and by deriving a probabilistic bound on the preservation of Euclidean structure.
Finally, Section \ref{sec:sf-rl} draws together all the results by discussing the use of sparse filtering as a representation learning algorithm.

\subsection{Information-Theoretic Aspects of Sparse Filtering \label{sec:info-theory}}

With reference to the information-theoretic description of unsupervised learning presented in Section \ref{ITL}, the aim of sparse filtering seems to be a pure optimization of the informativeness principle. Indeed, sparse filtering algorithm explicitly maximizes a property of the learned distribution (related to the informativeness principle), but it makes no reference to the problem of preserving information in the original distribution (related to the infomax principle); its loss function seems to be concerned only with the second term in Equation \ref{eq:ITL formula} and to disregard the first term.

However, based on our information-theoretic understanding of unsupervised learning we argue that, actually, sparse filtering must, in some way, take into account the infomax principle. In the following, we  demonstrate the following thesis:
\begin{quotation}
	\emph{Sparse filtering does satisfy the informativeness principle
		through the maximization of the proxy of sparsity and it satisfies
		the infomax principle through the constraint of preservation of the
		structure of cosine neighborhoodness of the data.}
\end{quotation}

\subsection{Informativeness Principle in Sparse Filtering \label{sec:informativeness}}

Showing that sparse filtering satisfies the informativeness principle
is straightforward. Since the explicit the minimization of the entropy $H[Z]$ is computationally hard,
the sparse filtering algorithm adopts the standard proxy of sparsity. 
Increasing the sparsity of the representations $\mathbf{Z}^{(i)}$ concentrates the mass of the pdf $p(Z)$ around zero; as the pdf $p(Z)$ gets closer to a Dirac delta function, its entropy is $H[Z]$ is minimized \citep{Principe2010,Hurley2009}. Using the formalism of \citet{Pastor2015}:
\begin{eqnarray*}
	-\ell_{1}\left(\mathbf{Z}\right)\uparrow & \equiv & H\left[Z\right]\downarrow,
\end{eqnarray*} 
that is, as the sparsity, measured by the negative $\ell_{1}$-norm of the learned representations $\mathbf{Z}^{(i)}$ increases, so the entropy of the pdf $p(Z)$ decreases.

\subsection{Infomax Principle in Sparse Filtering \label{sec:infomax}}

Showing that sparse filtering satisfies the informativeness principle
is more challenging. By definition, as a feature distribution learning
algorithm, sparse filtering does not address the problem of modeling
the data distribution. However, by virtue of the fact that sparse
filtering works and its learned representations $\mathbf{Z}^{(i)}$
allow the achievement of state-of-the-art performance when learning
$p\left(Y\vert Z\right)$, it \emph{must} be that the algorithm preserves
information contained in the original representations $\mathbf{X}^{(i)}$.
\emph{If it were not so}, sparse filtering could simply solve its
optimization problem by mapping the original data matrix $\mathbf{X}$
onto a pre-computed sparse representation matrix $\bar{\mathbf{Z}}$, containing a constant $1$-sparse learned representation $\bar{\mathbf{Z}}^{(1)}$,
with a minimal computational complexity of $O(1)$. The matrix $\bar{\mathbf{Z}}$ would have maximal sparsity, and the associated pdf $p(Z)$ would be a Dirac delta function centered on $\bar{\mathbf{Z}}^{(1)}$ with minimal entropy.
However, if we were to use $\bar{\mathbf{Z}}$ to perform further supervised learning with respect to a vector of label $\mathbf{Y}$, the pre-computed learned representations $\mathbf{Z}^{(i)}=\bar{\mathbf{Z}}^{(1)}$ would be useless as they would provide no information about the labels because of the independence between the pre-computed representations and the given labels: $p(Y\vert Z)=p(Y)$.

Since sparse filtering does not try to explicitly model the distribution of the original data we hypothesize that it must implicitly preserve information about the pdf $p(X)$ through the proxy of the preservation of data structure. 
The geometric structure of the data in the original space
$\mathbb{R}^{O}$ constitutes a set of realizations of the random
variable $X$ through which we can estimate the pdf $p(X)$. Preserving
relationships of neighborhoodness (under a given metric) allows us
to preserve information conveyed by the pdf $p(X)$: regions of high
density and low density in the domain of $p(X)$ can be
maintained by preserving relationships of neighborhoodness in the domain of $p(Z)$. Thus, preservation
of the geometric structure under a chosen metric may act as a proxy
for the maximization of mutual information $I[X;Z]$.

\subsection{Non-preservation of Euclidean Distance \label{sec:no-euclid}} 

The preservation of absolute or relative distances under the Euclidean metric is the most common way to preserve the structure of the data.
However, it can be easily ruled out that sparse filtering preserves this type of structure.

\begin{proposition} \label{non-preservation-euclidean}
	Let $\left\{ \mathbf{X}^{(i)}\in\mathbb{R}^{O}\right\} _{i=1}^{N}$ be a set of points in the original space $\mathbb{R}^{O}$. Then, the transformations from A1 to A4 do not preserve the structure of the data described by the Euclidean metric.
\end{proposition}

\textbf{Proof Sketch.} This proposition is proved by counterexample showing that there is at least a case for which the transformations from A1 to A4 do not preserve the Euclidean distance. The full proof is available in appendix \ref{app:non-preservation-euclidean}. $\blacksquare$


\subsection{Preservation of Collinearity \label{sec:pres-coll}}

Having ascertained that sparse filtering can not preserve the data
structure defined by the Euclidean metric, we investigate other properties
of the algorithm that may lead us to discover the preservation of
alternative data structures. A first relevant observation is that
sparse filtering preserves collinearity \revvv{I spelled out the definition of collinearity}{of point lying on the same line passing through the origin} of the space $\mathbb{R}^{O}$.

\begin{theorem}\label{collinearity}  Let $\mathbf{X}^{(1)},\mathbf{X}^{(2)}\in\mathbb{R}^{O}$ be collinear
points in the original space $\mathbb{R}^{O}$. Then, the outputs
of transformations from A1 to A4, that is $f_{A1:A4}\left(\mathbf{X}^{(1)}\right)$,  $f_{A1:A4}\left(\mathbf{X}^{(2)}\right)\in\mathbb{R}^{L}$,
are collinear.
\end{theorem}

Before proving this theorem, we present a set of auxiliary lemmas. \revvv{As these proofs are not very relevant I moved them to the appendix. This should make the reading more flowing}{The proofs of these lemmas are elementary and they can be found} in Appendices \ref{app:lem1}-\ref{app:lem4}.

\begin{lemma} \label{lem1} Let us consider $\mathbf{u},\mathbf{v}\in\mathbb{R}^{O}$, two generic collinear vectors, and let $f:\mathbb{\mathbb{R}}^{O}\rightarrow\mathbb{R}^{L}$	be a linear transformation defined as $f(\mathbf{u})=\mathbf{Wu}$, where $\mathbf{W}$ is the matrix associated with the linear transformation. Then, $f\left(\mathbf{u}\right),f(\mathbf{v})\in\mathbb{R}^{L}$ are also collinear.
\end{lemma}

\begin{lemma} \label{lem2} Let us consider $\mathbf{u},\mathbf{v}\in\mathbb{R}^{L}$, two generic collinear vectors, and let $f:\mathbb{\mathbb{R}}^{L}\rightarrow\mathbb{R}^{L}$ be the element-wise absolute-value function $f(\mathbf{u})=\left|\mathbf{u}\right|=\left[\left|u_{j}\right|\right]$. Then $f\left(\mathbf{u}\right),f(\mathbf{v})\in\mathbb{R}^{L}$ 	are also collinear.
\end{lemma}

\begin{lemma} \label{lem3} Let us consider $\mathbf{u},\mathbf{v}\in\mathbb{R}^{L}$, two collinear vectors whose components are all strictly positive\footnote{Notice that we can safely make the assumption of strict positivity in sparse filtering since $\mathbf{u}$ and $\mathbf{v}$ are the output of an absolute-value function implemented as $f(x)=\sqrt{x^{2}+\epsilon}$.}, and let $f:\mathbb{\mathbb{R}}^{L}\rightarrow\mathbb{R}^{L}$ be the $\ell_{2}$-normalization along the features. Then $f\left(\mathbf{u}\right),f(\mathbf{v})\in\mathbb{R}^{L}$ are also collinear.
\end{lemma}

\begin{lemma} \label{lem4} Let us consider $\mathbf{u}\in\mathbb{R}^{L}$, a vector whose components are all strictly positive \footnote{Notice that we can safely make the assumption of strict positivity in sparse filtering since $\mathbf{u}$ is the output of the normalization along the feature which preserves the positivity.}, and let $f:\mathbb{\mathbb{R}}^{L}\rightarrow\mathbb{R}^{L}$ be the $\ell_{2}$-normalization along the samples. Then $f\left(\mathbf{u}\right)\in\mathbb{R}^{L}$ have the same angular coordinates as $\mathbf{u}$. 
\end{lemma}

Using these lemmas, we can prove Theorem \ref{collinearity}.\\

\textbf{Proof of Theorem \ref{collinearity}.} To prove that the transformations from A1 to A4 preserve collinearity
it is necessary to prove that all transformations preserve collinearity.

Concerning transformation A1, by Lemma \ref{lem1}, linear transformations preserve collinearity. Concerning transformation A2, by Lemma \ref{lem2}, absolute-value function preserves collinearity; indeed, it rigidly folds all the orthants on the first one. Concerning transformation A3, by Lemma \ref{lem3}, normalization along the features preserves collinearity; indeed, it acts simply as a rescaling of the
axes. Concerning transformation A4, by Lemma \ref{lem4}, normalization along the samples preserves angular distances in general, and, therefore, collinearity.

Since all the transformations from A1 to A4 preserve collinearity,
the overall transformation $f_{A1:A4}(\cdot)$ preserves collinearity. $\blacksquare$

\subsection{Homo-representation of Collinear Points \label{sec:homorepresentation-collinear}}

An immediate consequence of the previous result is the following
theorem which states that all the collinear points in the original representation
space are mapped to an identical representation. This result is significant
as it gives us a first understanding of the principle and the type
of metric that sparse filtering uses to map original samples $\mathbf{X}^{(i)}$
onto their representations $\mathbf{Z}^{(i)}$.

\begin{theorem} \label{thm-homo-representation-collinear-point} Let $\mathbf{X}^{(1)}\in\mathbb{R}^{O}$ be a point in the original
space $\mathbb{R}^{O}$. Then there is a set of infinite points $\mathbf{X}^{(i)}\in\mathbb{R}^{O}$
such that $f_{A1:A4}\left(\mathbf{X}^{(1)}\right)=f_{A1:A4}\left(\mathbf{X}^{(i)}\right)$.
The set of the points collinear with $\mathbf{X}^{(1)}$ is included in this set.
\end{theorem}

\textbf{Proof.} 
Let us consider a point $\mathbf{X}^{(1)}$ and a generic collinear
point $\mathbf{X}^{(2)}=k\mathbf{X}^{(1)}$, $k\neq0$. Let us apply
the transformation $f_{A1:A4}$ to the points $\mathbf{X}^{(1)}$
and $\mathbf{X}^{(2)}$:
\[
\begin{array}{ccc}
f_{A1}\left(\mathbf{X}^{(1)}\right)=\mathbf{W}\mathbf{X}^{(1)} &  & f_{A1}\left(\mathbf{X}^{(2)}\right)=k\mathbf{W}\mathbf{X}^{(1)}\\
f_{A2}\left(\mathbf{W}\mathbf{X}^{(1)}\right)=\left|\mathbf{W}\mathbf{X}^{(1)}\right| &  & f_{A2}\left(k\mathbf{W}\mathbf{X}^{(1)}\right)=k\left|\mathbf{W}\mathbf{X}^{(1)}\right|\\
f_{A3}\left(\left|\mathbf{W}\mathbf{X}^{(1)}\right|\right)=\mathbf{c}\circ\left|\mathbf{W}\mathbf{X}^{(1)}\right| &  & f_{A3}\left(k\left|\mathbf{W}\mathbf{X}^{(1)}\right|\right)=k\left(\mathbf{c}\circ\left|\mathbf{W}\mathbf{X}^{(1)}\right|\right)\\
f_{A4}\left(\mathbf{c}\circ\left|\mathbf{W}\mathbf{X}^{(1)}\right|\right)=\frac{\mathbf{c}\circ\left|\mathbf{W}\mathbf{X}^{(1)}\right|}{\ell_{2}\left(\mathbf{X}^{(1)}\right)} &  & f_{A4}\left(k\left(\mathbf{c}\circ\left|\mathbf{W}\mathbf{X}^{(1)}\right|\right)\right)=\frac{k\left(\mathbf{c}\circ\left|\mathbf{W}\mathbf{X}^{(1)}\right|\right)}{\ell_{2}\left(\mathbf{X}^{(2)}\right)}
\end{array}
\]
where $\mathbf{c}$ is a vector of normalizing constants and $\circ$
is the element-wise product (as in Lemma \ref{lem3}). Now, since $\ell_{2}\left(\mathbf{X}^{(2)}\right)=k\ell_{2}\left(\mathbf{X}^{(1)}\right)$,
it follows:
\[
\frac{k\left(\mathbf{c}\circ\left|\mathbf{W}\mathbf{X}^{(1)}\right|\right)}{\ell_{2}\left(\mathbf{X}^{(2)}\right)}=\frac{k\left(\mathbf{c}\circ\left|\mathbf{W}\mathbf{X}^{(1)}\right|\right)}{k\ell_{2}\left(\mathbf{X}^{(1)}\right)}=f_{A4}\left(\mathbf{c}\circ\left|\mathbf{W}\mathbf{X}^{(1)}\right|\right).
\]
Thus, $\mathbf{X}^{(1)}$ and any collinear point $\mathbf{X}^{(2)}$
are mapped onto the same representation $f_{A1:A4}\left(\mathbf{X}^{(1)}\right)$. $\blacksquare$

\subsection{Homo-representation of Points with Same Moduli \label{sec:homorepresentation-moduli}}

A further analysis of sparse filtering reveals that not only collinear
points are mapped to the same representation, but also points in the
learned representation space having the same moduli (that is, the same
absolute value for their components) are mapped to identical representations.
Again, this result is relevant since it sheds light on the type of
structure preserved by sparse filtering.

\begin{theorem} \label{thm-homo-representation-same-moduli} Let $\mathbf{F}^{(1)}\in\mathbb{R}^{L}$ be a point in the codomain of the linear map defined by the matrix $\mathbf{W}$. It holds that for $\mathbf{F}^{(1)}$ strictly in the first orthant, there are at least $2^{L}$ points $\mathbf{F}^{(i)}\in\mathbb{R}^{L}$
such that $f_{A2:A4}\left(\mathbf{F}^{(1)}\right)=f_{A2:A4}\left(\mathbf{F}^{(i)}\right)$.
\end{theorem}

\textbf{Proof.} By definition, $\mathbf{F}^{(1)}_j > 0$, $\forall j$, $1 \leq j \leq L$. It follows that $f_{A2}(\mathbf{F}^{(1)})=\mathbf{F}^{(1)}$, as the application of the absolute-value maps $\mathbf{F}^{(1)}$ to itself.

However, all the vectors $\mathbf{F}^{(i)}$ such that $\mathbf{F}^{(i)}_j = \pm \mathbf{F}^{(1)}_j $ are mapped to $\mathbf{F}^{(1)}$ by the absolute-value $f_{A2}(\cdot)$. By combinatorial analysis, there are $2^{L}$ possible
ways of picking the values of $\mathbf{F}^{(i)}$, thus defining $2^{L}$
points in $\mathbb{R}^{L}$ that are mapped to the same value $\mathbf{F}^{(1)}$. Since all the points $\mathbf{F}^{(i)}$ are mapped to the same point $\mathbf{F}^{(1)}$ at the end of step A2,
the application of the remaining deterministic functions will map them to the same representation, $f_{A2:A4}\left(\mathbf{F}^{(1)}\right)=f_{A2:A4}\left(\mathbf{F}^{(i)}\right)$.
$\blacksquare$

\subsection{Preservation of Cosine Neighborhoodness \label{sec:pres-cosine-neigh}}

In Theorem \ref{thm-homo-representation-collinear-point} we have shown that sparse filtering maps points having the same angles to the same representation. However, this property is not sufficient to preserve any complex structure. Here we further prove that sparse filtering maps points having a small cosine distance in the original space onto point having small Euclidean distance in the representation space.

\begin{theorem} \label{preservation-cosine-neigh}
	Let $\mathbf{X}^{(1)},\mathbf{X}^{(2)}\in\mathbb{R}^{O}$
	be two original data samples and let $\mathbf{Z}^{(1)},\mathbf{Z}^{(2)}\in\mathbb{R}^{L}$
	be their representations computed by sparse filtering. If the cosine
	distance between the original samples is arbitrarily small $D_{C}\left[\mathbf{X}^{(1)},\mathbf{X}^{(2)}\right]<\delta$,
	for $\delta>0$, then the Euclidean distance between the computed
	representations is arbitrarily small $D_{E}\left[\mathbf{Z}^{(1)},\mathbf{Z}^{(2)}\right]<\epsilon$,
	for $\epsilon>0$, and $\epsilon=L \cdot \left(\frac{k+\left|\sqrt{2\delta-\delta^{2}}\right|}{\ell_{2}\left(\tilde{\mathbf{F}}^{(2)}\right)}-\frac{1}{\ell_{2}\left(\tilde{\mathbf{F}}^{(1)}\right)}\right)$,
	where $k$ is a constant accounting for partial collinearity and $\ell_{2}\left(\tilde{\mathbf{F}}^{(i)}\right)$
	is the $\ell_{2}$-norm of the representations computed by sparse
	filtering after step A3.
\end{theorem}

\textbf{Proof Sketch.} We provide here a synthetic sketch of the proof; the full proof is available in appendix \ref{app:preservation-cosine-neigh}.

We prove this theorem with the following approach: at each step of sparse filtering, (i) we consider the displacement between the representation of the two points $\mathbf{X}^{(1)}$ and $\mathbf{X}^{(2)}$; (ii) we upper bound the displacement.  

Before sparse filtering the displacement vector $\bar{\mathbf{X}}$ between $\mathbf{X}^{(1)}$ and $\mathbf{X}^{(2)}$ is:
\begin{equation}
\bar{\mathbf{X}}=(k-1)\mathbf{X}^{(1)}+\mathbf{B}
\end{equation}
where $k \in \mathbb{R}$, $k \neq 0$, is a constant accounting for partial collinearity and $\mathbf{B}$ is a bias vector. Knowing that the cosine distance between the samples is bounded by $D_{C}\left[\mathbf{X}^{(1)},\mathbf{X}^{(2)}\right]<\delta$, the displacement can be upper bounded component-wise as:
\begin{equation}
\bar{\mathbf{X}}_{j} \leq \mathbf{X}_{j}^{(1)}\left(k-1+\sqrt{2\delta-\delta^{2}}\right).
\end{equation}

After steps A1 and A2, the new displacement is:
\begin{equation}
\bar{\mathbf{F}}_{l}=\left(k-1\right)\mathbf{F}_{l}^{(1)}\pm\left|\mathbf{W}\mathbf{B}\right|_{l},
\end{equation}
whose upper bound is:
\begin{equation}
\bar{\mathbf{F}}_{l} \leq
\left(k-1+\left|\sqrt{2\delta-\delta^{2}}\right|\right)\left|\sum_{j=1}^{O}\mathbf{W}^{(l)}\mathbf{X}_{j}^{(1)}\right|.
\end{equation}

After the normalization along the rows in step A3, the displacement is reduced to:
\begin{equation}
\bar{\mathbf{\tilde{F}}}_{l}=\left(k-1\right)\tilde{\mathbf{F}}_{l}^{(1)}+\frac{\left|\mathbf{W}\mathbf{B}\right|_{l}}{c_{l}},
\end{equation}
where $\left\{ c_{l}\right\} _{l=1}^{L}$, $c\in\mathbb{R}$ are constant accounting for feature-dependent sums across the
$N$ samples. Consequently the new upper bound is simply:
\begin{equation}
\bar{\mathbf{\tilde{F}}}_{l} \leq
\frac{1}{c_{l}}\bar{\mathbf{F}}_{l}
\end{equation}

Finally, after step A4, the new displacement is:
\begin{equation}
\bar{\mathbf{Z}}_{l}=\left(k\frac{\ell_{2}\left(\tilde{\mathbf{F}}^{(1)}\right)}{\ell_{2}\left(\tilde{\mathbf{F}}^{(2)}\right)}-1\right)\mathbf{Z}_{l}^{(1)}+\frac{\left|\mathbf{W}\mathbf{B}\right|_{l}}{c_{l}\ell_{2}\left(\tilde{\mathbf{F}}^{(2)}\right)},
\end{equation}
which can be upper bounded as:
\begin{equation}
\bar{\mathbf{Z}}_{l} \leq
\frac{\left|\sum_{j=1}^{O}\mathbf{W}_{l}^{(j)}\mathbf{X}_{j}^{(1)}\right|}{c_{l}}\left(\frac{k+\left|\sqrt{2\delta-\delta^{2}}\right|}{\ell_{2}\left(\tilde{\mathbf{F}}^{(2)}\right)}-\frac{1}{\ell_{2}\left(\tilde{\mathbf{F}}^{(1)}\right)}\right).
\end{equation}
Therefore, the overall Euclidean distance between the representations $\mathbf{Z}^{(1)}$
and $\mathbf{Z}^{(2)}$ can be bounded by:
\begin{equation}
D_{E}\left[\mathbf{Z}^{(1)},\mathbf{Z}^{(2)}\right] \leq
L\cdot\left(\frac{k+\left|\sqrt{2\delta-\delta^{2}}\right|}{\ell_{2}\left(\tilde{\mathbf{F}}^{(2)}\right)}-\frac{1}{\ell_{2}\left(\tilde{\mathbf{F}}^{(1)}\right)}\right).
\end{equation}
Thus $\epsilon=L\cdot\left(\frac{k+\left|\sqrt{2\delta-\delta^{2}}\right|}{\ell_{2}\left(\tilde{\mathbf{F}}^{(2)}\right)}-\frac{1}{\ell_{2}\left(\tilde{\mathbf{F}}^{(1)}\right)}\right)$.
$\blacksquare$

Sparse filtering can then preserve cosine neighborhoodness by mapping points that have similar angular coordinates to representations that are close to each other under an Euclidean metric. However points that have large cosine distance in the original space will not necessarily be far in the representation space; this is a consequence
of the fact that transformation in sparse filtering preserve collinearity and cosine neighborhoodness, but not cosine metric in general.

\subsection{Basis and Basis Pursuit \label{sec:basis-pursuit}}

Let us now consider the space of the learned representations $\mathbb{R}^{L}$. This space is spanned by the canonical set of orthonormal bases $\left\{ \textbf{e}_i \right\}_{i=1}^L$, where $ \textbf{e}_1 = \left[\begin{array}{cccc}
1 & 0 & \ldots & 0 \end{array}\right]$, $\textbf{e}_2 = \left[\begin{array}{cccc}
0 & 1 & \ldots & 0 \end{array}\right]$, $\ldots$, $\textbf{e}_L = \left[\begin{array}{cccc}
0 & 0 & \ldots & 1 \end{array}\right] $.

Let $\mathsf{Z}$ be the set of vectors $\left\{ \mathbf{Z}^{(i)}\right\} _{i=1}^{N}$
produced by the sparse filtering algorithm through the steps A1 to A4. If we now consider the optimization
in step A5, it is easy to prove that the optimal set $\mathsf{Z}$
that minimizes the $\ell_{1}$-norm is given by a multi-set\footnote{We now explicitly refer to $\mathsf{Z}$ as a multi-set because the
optimal $\mathsf{Z}$ may contain repeated orthonormal bases of $\mathbb{R}^{L}$
in case $N>L$.} of the orthonormal bases of $\mathbb{R}^{L}$.

\begin{proposition} \label{basis-pursuit-proposition}
Let $\mathsf{Z}=\left\{ \mathbf{Z}^{(i)}\right\} _{i=1}^{N}$
be a set of vectors such that $\mathbf{Z}^{(i)}\in\mathbb{R}^{L}$
and $\sum_{j=1}^{L}\left(\mathbf{Z}_{j}^{(i)}\right)^{2}=1$. Then
an optimal set of vectors that solve the optimization problem $\displaystyle \min_{\mathbf{Z} \in \mathbb{R}^{L \times N}} $ $\sum_{i=1}^{N}\sum_{j=1}^{L}\mathbf{Z}_{j}^{(i)}$
is given by a multi-set of the orthonormal bases of $\mathbb{R}^{L}$.
\end{proposition}

\textbf{Proof Sketch.} This proposition is proved geometrically, following the proof given by \citep{Bishop2006} to show that the solutions to regularized least-squares optimization problems are sparse. The full proof is available in appendix \ref{app:basis-pursuit-proposition}. $\blacksquare$

Thus, the optimal solution for the sparse filtering algorithm is to map a set of original representations $\mathbf{X}^{(i)}\in\mathbb{R}^{O}$
onto the orthonormal bases of $\mathbb{R}^{L}$, as the bases $\left\{ \mathbf{e}_{i}\right\} _{i=1}^{L}$ have a minimal $\ell_{1}$-norm in $\mathbb{R}^{L}$ under the constraint of sparse filtering. 

Ideally, through gradient descent, sparse filtering progressively pushes all the learned representations $\mathbf{Z}^{(i)}\in\mathbb{R}^{L}$ towards the orthonormal bases of $\mathbb{R}^{L}$. However, in general, notice that sparse filtering is not guaranteed to find  an optimal solution and it may settle into a local minimum, where the original representations $\mathbf{X}^{(i)}$ are  mapped onto $k$-sparse ($k>1$) representations in $\mathbb{R}^{L}$. The quality of the solution depends on the original data set $\mathsf{X}$, on the dimensionality of the learned space $L$, and on the random initialization of the weight matrix $\mathbf{W}$.

\revvv{This paragraph summarizes the whole section 3.11}{}
Understanding the dynamics of sparse filtering in terms of bases and pursuit of bases naturally prompts a comparison with other sparse learning techniques used in signal processing and machine learning. 
\emph{Basis pursuit} \citep{Chen2001} defines a similar $\ell_{1}$-minimization problem, but it considers only a constraint given by a linear transformation, while sparse filtering transforms the data through non-linear transformations. 
\emph{Matching pursuit} \citep{Mallat1993} aims at learning a sparse representation; differently from sparse filtering which operates on all the data in parallel, matching pursuit works by finding a linear decomposition in an iterative way by selecting one basis at each iteration.  
Dictionary learning algorithms, such as the \emph{method of optimal directions} \citep{Engan1999} or \emph{k-Singular Value Decomposition} \citep{Aharon2006}, try to learn a dictionary and a sparse representation at the same time; however, they typically alternate between updating the dictionary and the sparse representation, while sparse filtering explicitly optimizes only the sparsity of the learned representation.

\subsection{Representation Filters \label{sec:repr-filters}}

\revvv{This section has been significantly reduced removing the discussion about the properties of representation filters}
The idea of orthonormal bases and pursuit of these bases allows us to introduce a last conceptual tool that gives us a better insight into the dynamics of sparse filtering. 

From Theorem \ref{thm-homo-representation-collinear-point} we learned that sparse filtering identifies sets of collinear points in the original space to be mapped onto bases; from Theorem \ref{thm-homo-representation-same-moduli} we can deduce that there must a symmetric structure around lines of collinear points; from Theorem \ref{preservation-cosine-neigh} we learned that cosine neighborhoodness is translated into Euclidean neighborhoodness. Putting together these results,
we can infer that sparse filtering defines precise maps in the
original representation space $\mathbb{R}^{O}$  in the form of \emph{representation filters}: 

\textbf{Definition (Representation Filter).} A \emph{representation filter} $R^{\mathbf{e}_j}$ is a function
$R^{\mathbf{e}_j}:\,\mathbb{R}^{O}\rightarrow\mathbb{R}_{\geq0}$
mapping points in the original representation space $\mathbb{R}^{O}$
to their Euclidean distance from the basis $\mathbf{e}_j$. 

Plotting a representation filter $R^{\mathbf{e}_j}$ in the original representation space $\mathbb{R}^{O}$ defines a region of space having a hyper-conical shape, such that all the points on the line of its height are mapped onto the basis $\textbf{e}_j$, and all the points in the neighborhood defined by its volume are mapped into the neighborhood of the basis $\textbf{e}_j$.
Moreover, given a point $\mathbf{X}^{(i)}\in\mathbb{R}^{O}$, we say
that the representation filter $R_{\mathbf{X}^{(i)}}^{\mathbf{e}_j}$
is centered on $\mathbf{X}^{(i)}$ if $R_{\mathbf{X}^{(i)}}^{\mathbf{e}_j} \left(\mathbf{X}^{(i)} \right) = 0 $, that is, the point $\mathbf{X}^{(i)}$
lies on the line of the height of the representation cone defined by $R_{\mathbf{X}^{(i)}}^{\mathbf{e}_j}$.

The optimization process of sparse filtering can be interpreted as the \emph{search for an optimal location of the representation filters}: hyper-conical representation filter are rotated in a continuous way in the original representation space during learning, until their placement provides an optimal solution in terms of sparsity of the learned representations. Inspecting the representation filters can provide a way to assess the progress of learning in sparse filtering.

\subsection{Non-preservation of Cosine Neighborhoodness in Alternative Implementations of Sparse Filtering \label{sec:alternative-non-linearities}}

The choice of the absolute-value non-linearity in step A2 of sparse filtering is crucial for guaranteeing the preservation of cosine neighborhoodness. 
\citet{Ngiam2011} suggest that this non-linearity may be substituted by other non-linear functions; for instance, standard non-linear functions from the neural networks literature, such as the sigmoid non-linearity or the rectified linear unit (ReLU), may be used. 
Despite this possibility, all the implementations of sparse filtering so far have relied on the absolute-value non-linearity.
An unpublished technical report by Thaler\footnote{\url{https://www.kaggle.com/c/challenges-in-representation-learning-the-black-box-learning-challenge/forums/t/4717/1st-place-entry}} states that sparse filtering with alternative non-linearities (ReLU and quadratic non-linearity) does not perform as well as the absolute-value non-linearity, but does not clarify the reasons of this failure. For plain empirical reasons, the absolute-value has always been recommended as the best non-linearity for sparse filtering.

One theoretical reason for the limited success of alternative implementations of sparse filtering is the fact that other non-linearities can not provide strong guarantees on preservation of relevant data structures. 
If we replace the absolute-value non-linearity with another non-linearity, such as sigmoid or ReLU function, we lose the property of preservation of structure guaranteed by standard sparse filtering, \revvv{Removed **immediate**}{} as it is proved by the following two propositions.

\begin{proposition} \label{non-preservation-sigm}
	Let us consider the sparse filtering algorithm implemented using a sigmoid non-linearity $\sigma(x)=\frac{1}{1+e^{-x}}$. 
	Let $\left\{ \mathbf{X}^{(i)}\in\mathbb{R}^{O}\right\} _{i=1}^{N}$ be a set of points in the original space $\mathbb{R}^{O}$. 
	Then, the transformations A1, A2*, A3 and A4, where A2* is the sigmoid non-linearity, do not preserve the structure of the data described neither by the Euclidean metric nor by the cosine metric.
\end{proposition}

\textbf{Proof Sketch.} This proposition is proved by counterexample. The full proof is available in appendix \ref{app:non-preservation-sigm}. $\blacksquare$

\begin{proposition} \label{non-preservation-relu}
	Let us consider the sparse filtering algorithm implemented using a soft ReLU non-linearity $ReLU(x)=\max\left(\epsilon,x\right)$, where $\epsilon$ is 	a small negligible value (for instance, $\epsilon=10^{-8}$). 
	Let $\left\{ \mathbf{X}^{(i)}\in\mathbb{R}^{O}\right\} _{i=1}^{N}$ be a set of points in the original space $\mathbb{R}^{O}$. 
	Then, the transformations A1, A2*, A3 and A4, where A2* is the ReLU non-linearity, do not preserve the structure of the data described neither by the Euclidean metric nor by the cosine metric.
\end{proposition}

\textbf{Proof Sketch.} This proposition is proved by counterexample. The full proof is available in appendix \ref{app:non-preservation-relu}. $\blacksquare$

The non-preservation of the Euclidean metric is not surprising and it is due to the fact that the normalization in step A4 does not preserve Euclidean distances. The non-preservation of cosine distances, collinearity or cosine neighborhoodness is instead due to the sigmoid and ReLU non-linearities in step A2, since these non-linearities do not determine a folding of the space in the same way of the absolute-value function. 

Despite this result, it may still be possible to implement sparse filtering with alternative non-linearities in order to preserve other types of structures. It is important that the preservation properties of alternative non-linearities agree with the structure preserved by the other steps of sparse filtering (A1, A3, A4). 
 From a theoretical perspective, the absolute-value non-linearity is a good choice for the sparse filtering algorithm, in that it preserves the common property of collinearity which is also preserved by all the other steps of the algorithm, therefore guaranteeing the preservation of the overall structure defined by cosine neighborhoodness.

\subsection{Bounds on Probability of Preserving Euclidean Neighborhoodness \label{sec:pres-euclidean}}

Interestingly, despite the fact that sparse filtering can not guarantee the preservation of the Euclidean metric, it is still possible to determine probabilistic bounds on the preservation of the Euclidean neighborhoodness under very simplified assumptions  on the dimensionality of the original space $\mathbb{R}^{O}$ and the region of space within which the samples $\mathbf{X}^{(i)}$ may be drawn.

\begin{theorem} \label{bounds-for-preserving}
	Let $\mathbf{X}^{(1)}\in\mathbb{R}^{O}$ be a point
	in the original space $\mathbb{R}^{O}$ and let $R_{\mathbf{X}^{(1)}}^{\mathbf{e}_{k}}$
	be a representation filter centered on $\mathbf{X}^{(1)}$, that is,
	$R_{\mathbf{X}^{(1)}}^{\mathbf{e}_{k}}\left(\mathbf{X}^{(1)}\right)=0$.
	Let us now consider a point $\mathbf{X}^{(2)}\in\mathbb{R}^{O}$ within
	the same representation cone, that is, a point such that $R_{\mathbf{X}^{(1)}}^{\mathbf{e}_{k}}\left(\mathbf{X}^{(2)}\right)\leq\epsilon$
	for an arbitrarily small $\epsilon\in\mathbb{R}$, $\epsilon>0$.
	
	Let us assume that: (i) points $\mathbf{X}^{(i)}$ distribute in a
	limited region of space bounded by $M$; and, (ii) points $\mathbf{X}^{(i)}$
	distribute uniformly in this limited region of space.
	
	Then, given that $R_{\mathbf{X}^{(1)}}^{\mathbf{e}_{k}}\left(\mathbf{X}^{(2)}\right)\leq\epsilon$, it follows:
	\[
	\frac{O\delta}{\left(\frac{M}{m}\right)^{O-1}}\cdot\frac{\Gamma\left(\frac{O+1}{2}\right)}{\Gamma\left(\frac{O+2}{2}\right)}\leq
	P\left(D_{E}\left[\mathbf{Z}^{(1)},\mathbf{Z}^{(2)}\right]\leq\delta\right)
	\leq\frac{O\delta}{m}\cdot\frac{\Gamma\left(\frac{O+1}{2}\right)}{\Gamma\left(\frac{O+2}{2}\right)},
	\]
	where $\delta\in\mathbb{R}$, $\delta>0$ defines the neighborhood
	of $\mathbf{X}^{(1)}$, $m$ is the distance of $\mathbf{X}^{(1)}$
	from the origin, and $\Gamma(\cdot)$ is the gamma function.
\end{theorem}

\textbf{Proof Sketch.} This proposition is proved geometrically, evaluating the limit of the ratio between the volume of a representation filter and the neighborhood of the a point when the dimensionality tends to infinity. The full proof is available in appendix \ref{app:bounds-for-preserving}. $\blacksquare$

Notice that this proof is based on two simplified assumptions. First,
the region of the original space in which a point $\mathbf{X}^{(i)}$
can fall is limited;  this assumption is reasonable because, practically,
the range of any feature is always bounded within a certain interval. Second, a
point $\mathbf{X}^{(i)}$ has a uniform probability of falling anywhere
within the area defined by the representation filter $R_{\mathbf{X}^{(1)}}^{\mathbf{e}_{k}}$;
this is clearly a simplified assumption because the pdf of the data
$p(X)$ may have a very irregular distribution within the area defined
by the representation filter $R_{\mathbf{X}^{(1)}}^{\mathbf{e}_{k}}$;
however,  assuming a uniform
distribution, which is a distribution that maximizes our uncertainty,
seems a reasonable choice.
If these two assumptions are accepted, approximate bounds can be computed
to evaluate the probability that sparse filtering will preserve relationship
of Euclidean neighborhoodness.

\subsection{Sparse Filtering for Representation Learning \label{sec:sf-rl}}

Given the above results, we may now interpret sparse filtering as
a soft clustering algorithm for representation learning.

Indeed, we may state that sparse filtering implicitly makes all the
assumptions made by traditional soft clustering algorithms (see Section \ref{sec:unsupervised_learning}): (i) it aims at discovering less noisy representations $ \textbf{Z}^{(i)} $ whose pdf $ p(Z) $ may automatically be closer to the true stochastic generating process with pdf $p\left(X^{*}\right)$; (ii) it expects the true pdf $p\left(X^{*}\right)$ to have a stronger correlation to the labels $\mathbf{Y}^{(i)}$;
(iii) it models the true pdf $p\left(X^{*}\right)$ with a mixture
model whose components are related to the bases $\{\mathbf{e}_j\}_{j=1}^L$; and, (iv) it relies on the cosine metric to evaluate relationships of neighborhoodness in the original space $\mathbb{R}^{O}$. 
From this perspective, we can interpret the dimensionality of the learned space as the number of clusters for soft clustering, the bases as the cluster centroids in a space described by the cosine metric, the pursuit of the bases as the sequential process of update of the location of the centroids, and the learned representations $\mathbf{Z}^{(i)}$ as the (stochastic)
degree of membership of the original data samples $\mathbf{X}^{(i)}$ to each cluster.

Given this interpretation, we can align and meaningfully compare sparse filtering with other soft clustering algorithms for representation learning that use different metrics. The choice of an appropriate metric is critical for a distance-based clustering algorithm \citep{Xing2003}, and it expresses our understanding on which spatial directions encode relevant changes \citep{Simard1998}. It is natural then to compare
sparse filtering with other standard algorithms which adopt the Euclidean metric to explain the data. 
 Preserving the relationships of neighborhoodness under the Euclidean metric means preserving the information conveyed by the pdf $p(X)$ in the representation space defined by the Cartesian product of the random variables $X_{1},X_{2},\ldots,X_{O}$, while preserving the relationships of neighborhoodness under the cosine metric means
preserving information conveyed by the pdf $p(X)$ in the representation space defined by the projection into polar (or hyper-spherical) coordinates of the random variables $X_{1},X_{2},\ldots,X_{O}$.
Choosing one metric instead of the other depends on our expectation whether the structure of the data with respect to a set of labels $p(Y\vert X)$ is better explained by an Euclidean structure or by a radial structure.

\section{Empirical Validation \label{sec:Experimental-Simulation}}

Based on the theoretical analysis provided in the previous section, we conduct a set of simulations aimed at verifying our theoretical results empirically. In order to make our results visualizable and easily understandable, we first conduct simple simulations in low dimensions; experiments in higher dimensions generalize our results but they do not add anything conceptually new to our conclusions. We further validate our theoretical findings with a number of benchmark data sets pertaining to real-world applications.

\subsection{Properties of Sparse Filtering}

First, we run simulations on elaborately designed toy data
sets in order to validate our basic understanding of sparse filtering.
These simulations aim at verifying: (i) the property of homo-representation
of collinear points (see Section \ref{sec:homorepresentation-collinear});
(ii) the usefulness of representation filters (see Section \ref{sec:repr-filters});
and, (iii) the dynamics of pursuit of bases (see Section \ref{sec:basis-pursuit}). 

We generate a random set of data $\mathsf{X}$ of three samples ($N=3$)
in two-dimensional space ($O=2$). Each point is generated using spherical
coordinates: the radial distance $\rho$ is sampled from a uniform
distribution $Unif\left(-5,5\right)$; the angular coordinate $\theta$
is set to $\frac{\pi}{3}$ for the first two points and sampled from
a uniform distribution $Unif\left(0,\pi\right)$ for the third point.
A sparse filtering module is trained on this data set in order to
learn a new representation of the data in two dimensions ($L=2$).
After training, we create a dense mesh of points $\mathbf{X'}$ in
the original representation space $\mathbb{R}^{O}$; we project each
point $\mathbf{X'}$ to its representation $\mathbf{Z'}$ in the learned representation
space $\mathbb{R}^{L}$, and we compute the distance from each basis
$\mathbf{e}_j$ in $\mathbb{R}^{L}$. The plot of each representation
cone is then shown as a two-dimensional contour plot in the original
space $\mathbb{R}^{O}$.

\begin{figure}
\begin{centering}
\includegraphics[scale=0.4]{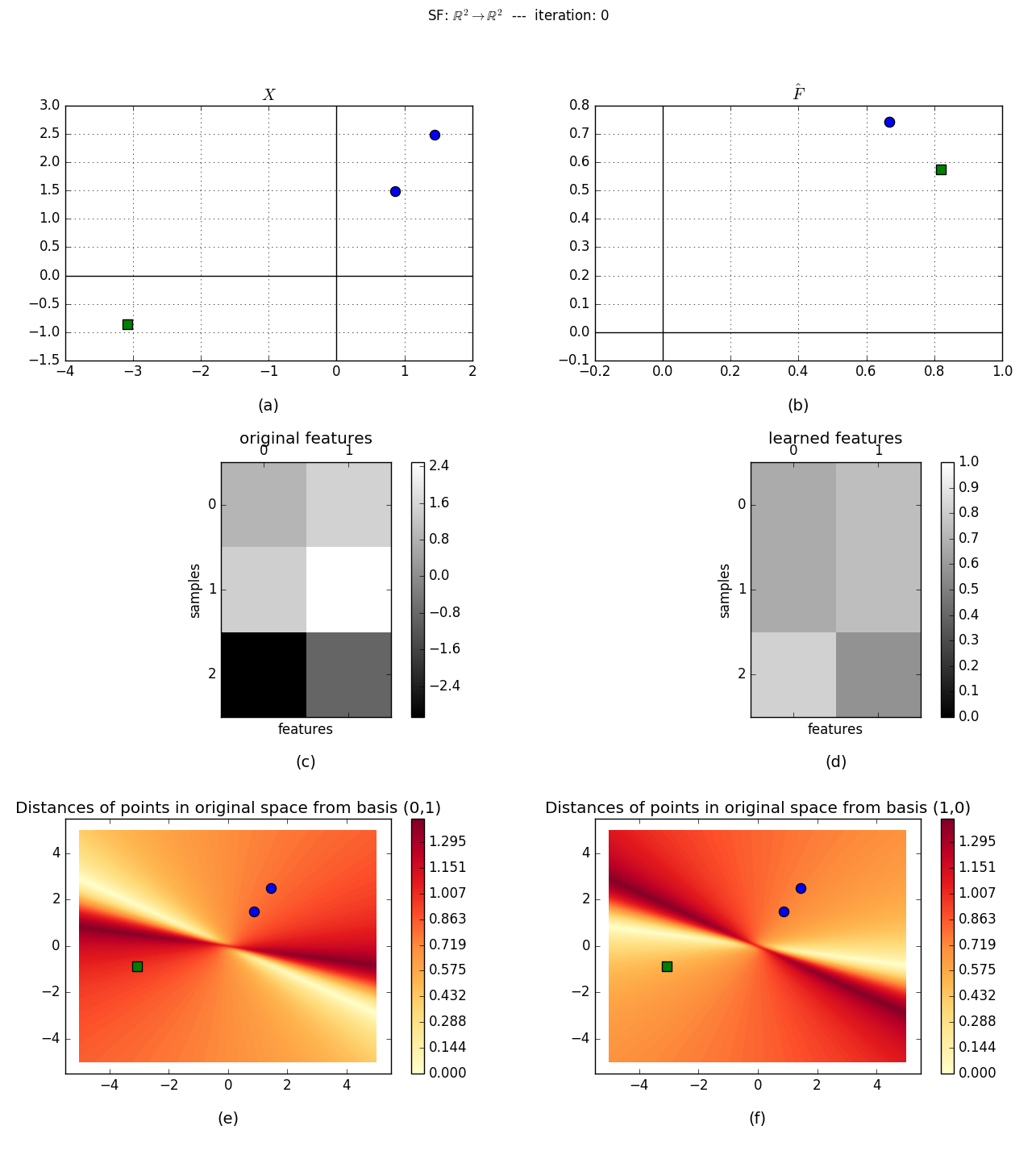}
\par\end{centering}

\caption{Experimental validation of the properties of sparse filtering (homo-representation
of collinear points, representation filters).\protect \\
Data is generated as explained in the text (the blue circle dots represent
collinear points). (a) Data in the original representation space $\mathbb{R}^{O}$;
(b) data in the learned representation space $\mathbb{R}^{L}$; (c)
matrix plot of the original data $\mathbf{X}$; (d) matrix plot of
the learned representations $\mathbf{Z}$; (e) plot of first representation
filter showing distances from the basis $\mathbf{e}_1=[0,1]^T$; (f)
plot of the second representation filter showing distances from the
basis $\mathbf{e}_2=[1,0]^T$. \label{fig:RC-9-1-begin}}
\end{figure}

Figure \ref{fig:RC-9-1-begin} shows the state of sparse filtering
before training. From the plots \ref{fig:RC-9-1-begin}(b) and \ref{fig:RC-9-1-begin}(d)
we can immediately verify the property of homo-representation of collinear
points; indeed, in the learned space $\mathbb{R}^{L}$ the collinear
points occupy the same location and their matrix representation is
the same. From the plots \ref{fig:RC-9-1-begin}(e) and \ref{fig:RC-9-1-begin}(f)
we can verify the existence of representation filters in the original
space $\mathbb{R}^{O}$ and appreciate how points in the original space are mapped onto bases of the learned space.  Notice that, at this point, after the
random initialization of the weight matrix $\mathbf{W}$, the quality
of the representations generated by the untrained sparse filtering module
is far from satisfactory.

\begin{figure}
\begin{centering}
\includegraphics[scale=0.4]{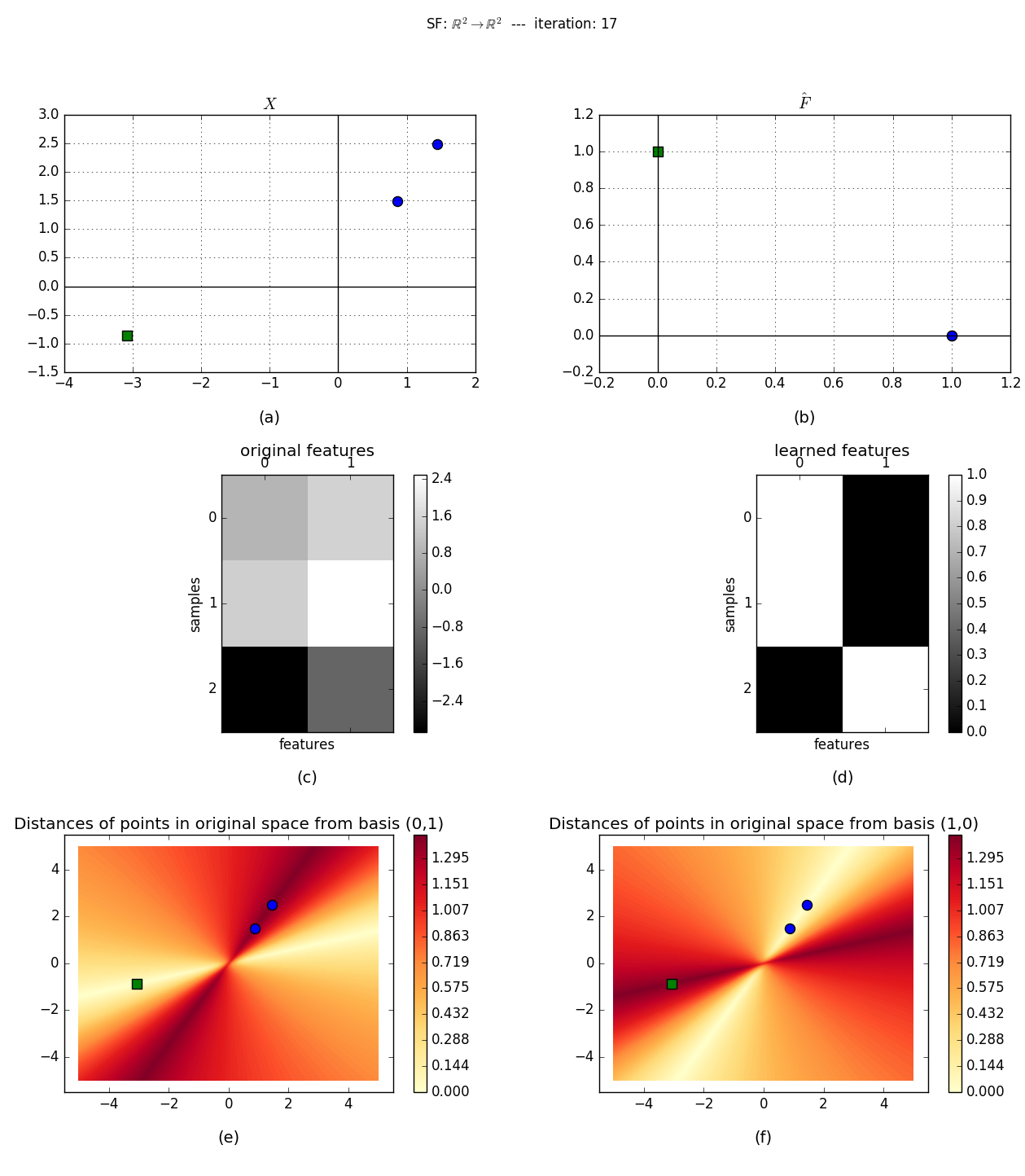}
\par\end{centering}

\caption{Experimental validation of the properties of sparse filtering (pursuit of bases).\protect \\
Data is generated as explained in the text. The meaning of the subplots
is the same as in Figure \ref{fig:RC-9-1-begin}. \label{fig:RC-9-1-end}}
\end{figure}

Figure \ref{fig:RC-9-1-end} shows the state of sparse filtering at
the end of training. From the plots \ref{fig:RC-9-1-end}(b) and
\ref{fig:RC-9-1-end}(d) we can see that the trained sparse filtering
module has found an optimal solution that maps all the points to
bases; as expected, the collinear points are mapped to the same basis,
while the third point is mapped to the remaining basis. From the plots
\ref{fig:RC-9-1-end}(e) and \ref{fig:RC-9-1-end}(f) we can validate
our intuition about the pursuit of bases; indeed, training corresponded to
a rotation of the representation filters in order to center them on
the available samples. 

\subsection{Preservation of Cosine Neighborhoodness \label{sub:Preservation-of-Cosine}}

Next, we run more simulations on other toy data sets in order to validate
the properties of data structure preservation in sparse filtering.
These simulations aim at verifying: (i) that sparse filtering preserves
a structure defined by cosine neighborhoodness (see Section \ref{sec:pres-cosine-neigh}); and, (ii) that the
absolute-value non-linearity is crucial in preserving structure and
substituting it with other non-linearities
negates this property (see Section \ref{sec:alternative-non-linearities}). 

We generated a random set of data $\mathbf{X}$ of nine samples ($N=9$)
in two-dimensional space ($O=2$). Each point is generated by randomly sampling its spherical coordinates.  The first three points have a radial distance $\rho \sim Unif\left(-2,0\right)$ and an
angular coordinate $\theta \sim Unif\left(\frac{\pi}{9}-\eta,\frac{\pi}{9}+\eta\right)$;
the following three points have a radial distance $\rho \sim Unif\left(0,3\right)$ and an angular coordinate
$\theta \sim Unif\left(\frac{2\pi}{9}-\eta,\frac{2\pi}{9}+\eta\right)$;
the last three points have a radial distance $\rho \sim Unif\left(2,4\right)$ and an angular coordinate
$\theta \sim Unif\left(\frac{4\pi}{9}-\eta,\frac{4\pi}{9}+\eta\right)$.
The parameter $\eta$ is meant to represent a form of noise and its value is set to $\eta=\frac{\pi}{45}$. In this way, we generate
three clusters of points, such that the cosine distances among the
points belonging to the same cluster are small, while the cosine distances
among points belonging to different clusters are large. Three implementations
of sparse filtering with different non-linearities (absolute-value, sigmoid, and ReLU\footnote{ReLU has been implemented in a soft version, like the absolute-value:
$softReLU(x)=\begin{cases}
x & if\,x>0\\
\epsilon & otherwise
\end{cases}$ for $\epsilon=10^{-8}$}) are used to learn a new representation of the data in two dimensions
($L=2$).

\begin{figure}
\begin{centering}
\includegraphics[scale=0.4]{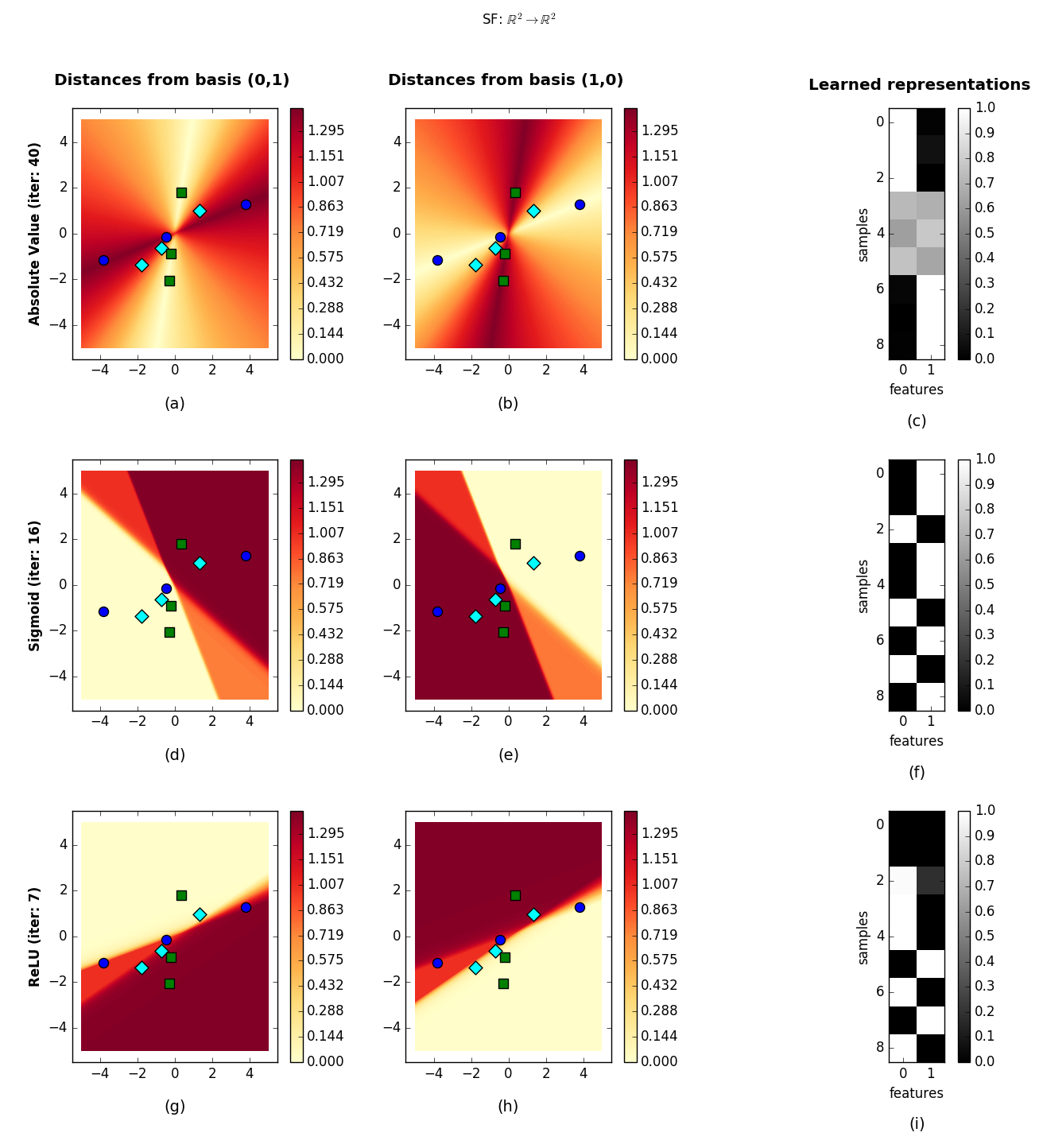}
\par\end{centering}

\caption{Experimental validation of the preservation of cosine neighborhoodness.\protect \\
Data is generated as explained in the text (first set of points as
blue circle dots, second set of points cyan diamond dots, third set of points as green square dots).
(a, d, g) Plot of the first representation filter showing distances
from the basis $\mathbf{e}_1=[0,1]^T$, respectively for the sparse
filtering with absolute-value, sigmoid, and ReLU non-linearity; (b,
e, h) plot of the second representation filter showing distances from
the basis $\mathbf{e}_2=[1,0]^T$, respectively for the sparse filtering
with absolute-value, sigmoid, and ReLU non-linearity; (c, f, i) matrix
plot of the learned representations $\mathbf{Z}$, respectively for
the sparse filtering with absolute-value, sigmoid, and ReLU non-linearity.
\label{fig:RC-10-1}}
\end{figure}

Figure \ref{fig:RC-10-1} shows the state of the modules of the three
implementations of sparse filtering at the end of the training. From
the plots \ref{fig:RC-10-1}(a)-\ref{fig:RC-10-1}(c) we
can immediately verify that sparse filtering with an absolute-value non-linearity preserves cosine neighborhoodness. The plots of representation filters show that points with similar angular coordinates fall
within the same representation filter. The matrix plot shows that points
with similar angular coordinates are projected onto very similar
representations; in other words, points that originally had a small
cosine distance are projected onto almost identical representations.
On the other hand, from plots \ref{fig:RC-10-1}(d)-\ref{fig:RC-10-1}(i)
we can easily see that sparse filtering with an alternative non-linearity
does not preserve cosine neighborhoodness. The plots of representation filters
show that the sigmoid and the ReLU non-linearity do not induce representation
cones, but, instead define large regions of the original space to
be mapped onto a basis;  several points are thus
indistinctly mapped onto a basis. The matrix plots show that the representations
computed by these alternative sparse filtering modules are not related
to the original cosine distances anymore; points originally belonging
to the same cluster are mapped to opposite representations, and, vice
versa, points originally belonging to different clusters are mapped
to identical representations.

\subsection{Sparse Filtering for Representation Learning \label{sub:Sparse-Filtering-for}}

In the following set of simulations, we compare sparse filtering against
another unsupervised algorithm in order to show under which conditions sparse filtering is a good
choice for processing data. These simulations aim at verifying the
following intuitive implication: if the structure of the
data with respect to a specific set of labels $p(Y \vert X)$ is better explained
by the cosine metric, then sparse filtering is likely to
be a good option for unsupervised learning. 

In our comparison, we measure sparse filtering against the soft $k$-means
algorithm \citep{MacKay2003}. We choose this algorithm for the following reason: (i) like
sparse filtering, the soft $k$-means algorithm is a soft clustering
algorithm producing sparse representations; (ii) the algorithm is
based on the Euclidean metric, thus providing a different interpretation
of the data from sparse filtering; and, (iii) $k$-means is a well-known and
easy-to-interpret algorithm (even if, analogous results may be obtained
with other algorithms, such as Gaussian mixture models or sparse auto-encoders).

To validate our hypothesis, we generate two data sets, $\mathbf{X}_{Euclid}$
and $\mathbf{X}_{cosine}$. The data set $\mathbf{X}_{Euclid}$ contains
data where $p(Y \vert X)$ is explained by the Euclidean metric. It is composed of nine samples
($N=9$) in two dimensions ($O=2$) sampled from three multivariate normal distribution. The first three points are sampled
from $\mathcal{N}\left(\left[\begin{array}{c}
1\\
1
\end{array}\right],\left[\begin{array}{cc}
.05 & 0\\
0 & .05
\end{array}\right]\right)$; the second three points are sampled from  $\mathcal{N}\left(\left[\begin{array}{c}
2\\
-1
\end{array}\right],\left[\begin{array}{cc}
.05 & 0\\
0 & .05
\end{array}\right]\right)$; the last three points are sampled from  $\mathcal{N}\left(\left[\begin{array}{c}
-1\\
-1
\end{array}\right],\left[\begin{array}{cc}
.05 & 0\\
0 & .05
\end{array}\right]\right)$. The data set $\mathbf{X}_{cosine}$ contains data where $p(Y \vert X)$ is explained by the
cosine metric. The data is generated following the same protocol used
in the simulation in Section \ref{sub:Preservation-of-Cosine}. Sparse
filtering is used to learn a new representation of the data in three
dimensions ($L=3$).

\begin{figure}
\begin{centering}
\includegraphics[scale=0.4]{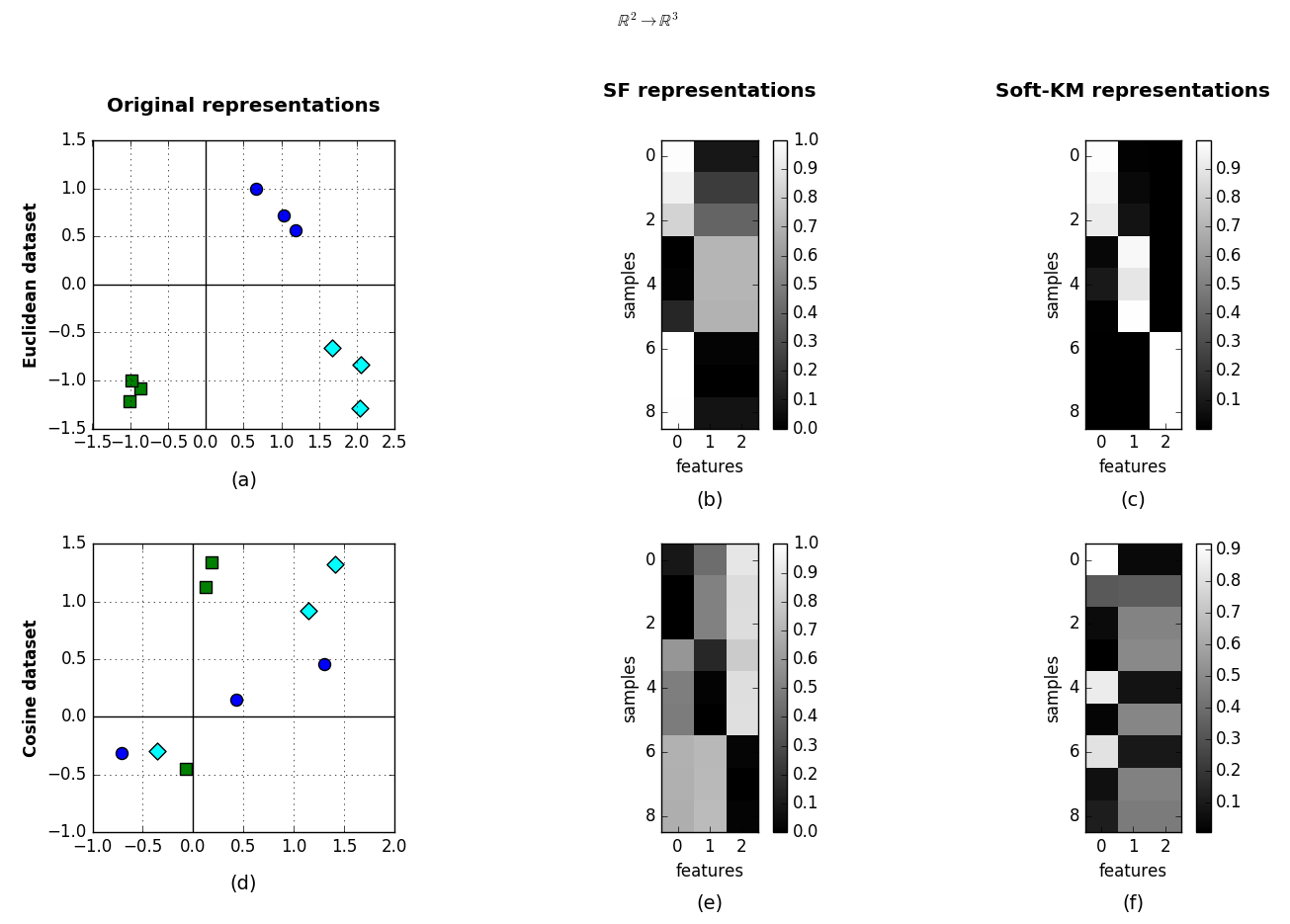}
\par\end{centering}

\caption{Representation data with Euclidean and cosine data structure.\protect \\
Data is generated as explained in the text (first set of points as
blue circle dots, second set of points cyan diamond dots, third set of points as green square dots).
(a, d) Samples in the original space; (b, e) matrix plot of the representations
learned by sparse filtering; (c, f) matrix plot of the representations
learned by soft $k$-means. \label{fig:tkm-2-end}}
\end{figure}

From Figure \ref{fig:tkm-2-end}, we can see that our understanding
of sparse filtering is correct: if $p(Y \vert X)$ is
better explained by the cosine metric, then sparse filtering produces
a good representation; otherwise, if $p(Y \vert X)$ is better explained
by the Euclidean metric, then it is reasonable to opt for a different unsupervised
learning algorithm, such as soft $k$-means. In the case of the data
set with Euclidean structure, plot \ref{fig:tkm-2-end}(b) shows that
sparse filtering is not able to preserve the identity of the generating
clusters, and indeed it maps samples from
the first and the third clusters onto the same representation (because of their collinearity);
instead, plot \ref{fig:tkm-2-end}(c) shows that soft $k$-means algorithm
maps points from different clusters to different representations.
In contrast, in the case of the data set with cosine structure, plot
\ref{fig:tkm-2-end}(e) shows that sparse filtering preserves the
identity of the generating clusters, while plot \ref{fig:tkm-2-end}(f)
shows that the soft $k$-means algorithm is unable to map samples from
different clusters onto consistent representations.

\subsection{Sparse Filtering on Real Data Sets}

In this last set of simulations we apply our discoveries about sparse
filtering to real-world data sets to further verify our results. Once
again, these experiments aim at validating the connection between
the radial structure of the data and the success of sparse filtering.
In the first simulation, we extend the result that we proved in Section
\ref{sub:Sparse-Filtering-for} for toy data sets to real data sets;
that is, we verify the direct implication: \emph{if} the
structure of the data with respect to a specific set of labels $p(Y \vert X)$ is
better explained by the cosine metric, \emph{then} sparse filtering
is likely to be a good option for unsupervised learning. In the second
simulation, we validate, instead, the reverse implication:
\emph{if} sparse filtering happens to be a good option for unsupervised
learning, \emph{then} the structure of the data with respect
to a specific set of labels $p(Y \vert X)$ is likely to be better explained by the
cosine metric.

When dealing with real data sets, it is very challenging
to assess the structure of the data. With few samples in low dimensions
and with the simplified assumption that all the data belonging to
a given class are generated by a single highly localized cluster, a simple visualization or a computation of distances is
enough to understand which metric is underlying the data.  
However, when we consider real
data sets, we have to deal with a large number of samples in high dimensions, and with the fact that samples belonging to
the same class may be generated by different clusters spread throughout
the space. In order to explore high-dimensional data in the original space before any normalization, we decided to rely on the
$k$-nearest neighbors algorithm (KNN). We implemented two versions
of KNN, one selecting $k$ neighbors according to the Euclidean distance
and one selecting $k$ neighbors according to the cosine distance\footnote{The KNN using cosine distance has been implemented relying on the
``trick'' that the cosine distance between vectors $\mathbf{u},\mathbf{v}$
is the same as the Euclidean metric on the $\ell_{2}$-normalized
vectors. Therefore, we perform an $\ell_{2}$-normalization of each
data sample and then we run KNN with Euclidean distance, re-using
off-the-shelf KNN code optimized for the Euclidean metric.}. If $p(Y \vert X)$ is
better explained by the Euclidean distance, we expect KNN with the
Euclidean metric to provide better results; alternatively, if $p(Y \vert X)$ is better explained
by the cosine distance, we expect KNN with the cosine metric to provide
better results. \\

\textbf{Berlin Emotional data set.} The Berlin Emotional (EMODB) data set is a well-known audio data set
in the emotion recognition community \citep{Burkhardt2005}; it contains
recordings of ten German actors expressing seven different types of
emotions. We opted for this data set to validate the direct implication
between data structure and effectiveness of sparse filtering for the
following reasons. (i) Samples in EMODB naturally lend themselves
to alternative labellings; the same data may be used
both for speaker recognition (using subject labels) and for emotion
recognition (using emotional labels). (ii)  The same set of Mel-frequency
cepstrum \citep{Childers1977} coefficient (MFCC) features may reasonably be used both for speaker recognition and for emotion recognition  \citep{Wu2010,Schuller2011a}.
Thus, the same features we can be used to explore the data under different
labeling.

We first explore the structure of the data with respect to the two different
labeling systems in order to evaluate whether the Euclidean distance
or the cosine distance better explains the structure of the data. We
run the KNN algorithm with different values of neighbors ($k=\{2,3,5,7,10,15,20,25,50,75,100\}$);
for each configuration of KNN, fifty simulations are executed; in
each simulation the data set is randomly partitioned into a training
data set (900 samples) and a test data set (311 samples); KNN is
then trained and tested using one of the two available metrics. 

After this analysis, we use both an Euclidean-based unsupervised learning
algorithm, Gaussian mixture model \citep{Bishop2006}, and a cosine-based
unsupervised learning algorithm, sparse filtering, to project the
data into an $L$-dimensional space. We opted for the Gaussian mixture
of models (GMM) algorithm because it is based on the Euclidean metric
and yields better performance than the soft $k$-means algorithm.
After processing the data, we then run a simple linear SVM classifier
on the processed data and we analyze how our observations on the structure
of the data relate with the actual classification performance. We
consider several values of dimensionality ($L=\{2,3,...,40\}$); for
each configuration, fifty simulations are executed; as before, in
each simulation the data set is randomly partitioned into a training
data set (900 samples) and in a test data set (311 samples). 

\begin{figure}
\begin{centering}
\includegraphics[scale=0.55]{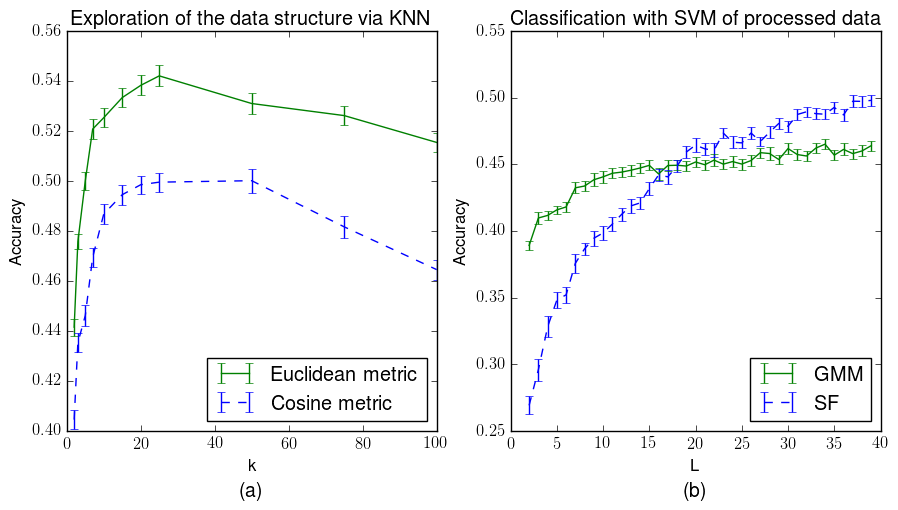}
\par\end{centering}

\caption{Analysis of the data structure and the classification of the EMODB
data set with respect to emotion labels.\protect \\
Classification is performed as explained in the text. (a) Exploration
of the data via KNN with Euclidean metric (green continuous line) and with cosine
metric (blue dashed line); (b) Classification using a
linear SVM after processing with a GMM algorithm (green line) and
with sparse filtering (blue line). The plot shows the average accuracy
and the standard error of SVM (over fifty simulations). \label{fig:EMODBe}}
\end{figure}

\begin{figure}
\begin{centering}
\includegraphics[scale=0.55]{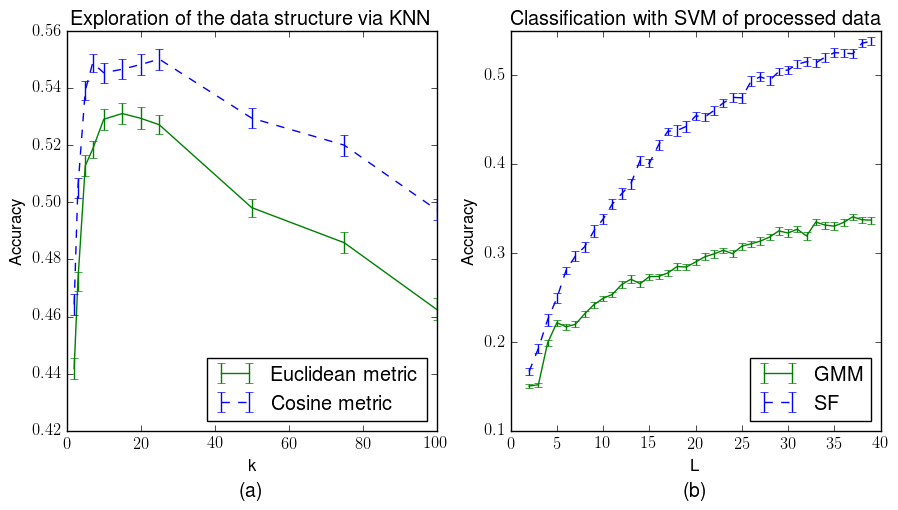}
\par\end{centering}

\caption{Analysis of the data structure and the classification of the EMODB
data set with respect to subject labels.\protect \\
The meaning of the subplots
is the same as in Figure \ref{fig:EMODBe}.\label{fig:EMODBs}}
\end{figure}

Figure \ref{fig:EMODBe}(a) shows that the structure of EMODB data with
respect to emotional labels is better explained by the Euclidean distance.
This result is further confirmed by the classification with the linear
SVM module in Figure \ref{fig:EMODBe}(b). Classification using the GMM-processed data with low learned
dimensionality ($L\leq15$) returns an accuracy that is significantly
better than using sparse filtering-processed data (Wilcoxon signed-rank
test, p-value $P=5\cdot10^{-85}$); however, in higher dimensions the
classification with sparse filtering-processed data approaches and
overtakes the accuracy obtained using GMM-processed data. In general,
in low dimensions, the Euclidean structure assumed by GMM explains
the data better; in high dimensions, sparse filtering provides
good results (most likely thanks to the property of sparsity), but
the gap between the accuracy provided by the two representations remains
limited. On the other hand, Figure \ref{fig:EMODBs}(a) shows that the
structure of EMODB data with respect to the speaker identity labels
is better explained by the cosine distance. This result is further confirmed
by the classification with the linear SVM module in Figure \ref{fig:EMODBs}(b). Classification using
the sparse filtering-processed data returns, for all learned dimensionality,
an accuracy that is significantly better than GMM-processed data (Wilcoxon
signed-rank test, p-value $P=4\cdot10^{-307}$). The assumption of the
cosine metric allows sparse filtering to explain the data much better,
as is evident from the large gap between the accuracy provided by the
two representations. 

These results confirm a connection between the radial structure
of the data with respect to a set of labels and the usefulness of
sparse filtering.\\

\textbf{Kaggle Black Box Learning Challenge data set.} The Kaggle Black Box Learning Challenge (KBBLC) data set is a visual
data set made up of obfuscated images of house numbers; the original
images are taken from the well-known Street View House Numbers (SVHN)
data set \citep{Netzer2011}. 
Each sample in the KBBLC data set contains
a single obfuscated digit and it is accompanied by a label specifying
the value of the digit. We opted to validate the reverse implication
between data structure and effectiveness of sparse filtering on this data set for the
following reasons. (i) Sparse filtering provided state-of-the-art
performance in the competitive KBBLC contest, thus showing that sparse
filtering was a particularly suitable choice for this data set. (ii)
The KBBLC data set is available with labels. During the challenge
the authors provided obfuscated data without labels; however, after
the challenge they revealed the original source of the data\footnote{\url{http://ufldl.stanford.edu/housenumbers/}}
and they released the code they used for obfuscation\footnote{\url{https://www.kaggle.com/c/challenges-in-representation-learning-the-black-box-learning-challenge/forums/t/5167/the-data}}.
Thanks to this information, we were able to retrieve a large amount
of data and obfuscate it, and thus recreate the original conditions
of the challenge. However, differently from the challenge, we retain
the labels in order to explore the structure of the data. (iii)
During the challenge, the original samples from the data sets were processed without undergoing operations of windowing or convolution. Since sparse filtering was directly applied to the
samples, we can analyze the structure of the samples straightforwardly.
This condition is not always true. If we consider other image data sets
on which sparse filtering provided good results, such as CIFAR-10 \citep{Krizhevsky2009}
or STL-10 \citep{Coates2011}, sparse filtering was not applied to the original samples
but to random patches extracted from the images; in this case, we
should not analyze the data structure of the original samples, but
the data structure of the patches. However, patches are not labeled,
which hinders our ability to carry out an analysis of the data structure.

In exploring the structure of the data (with respect to the digit
labels), we aim at evaluating whether the Euclidean distance or the cosine
distance better explains $p(Y \vert X)$. We run the KNN
with the same settings as in the previous experiment. In each simulation
a random subset of 10000 samples from the data set was selected and
then partitioned into a training data set (9000 samples) and a test
data set (1000 samples). KNN was then trained and tested using one
of the two available metrics. 

\begin{figure}
\begin{centering}
\includegraphics[scale=0.55]{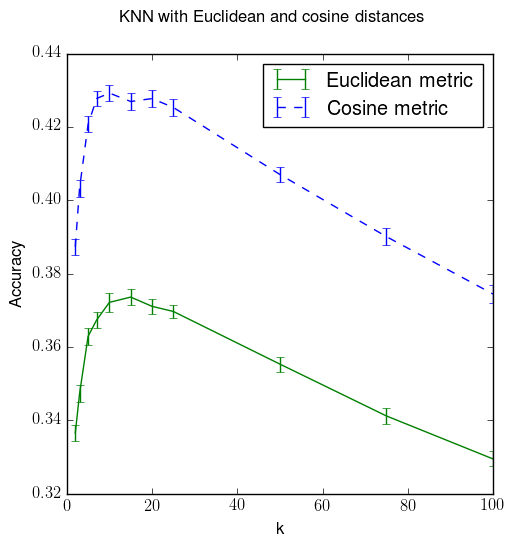}
\par\end{centering}

\caption{Analysis of the data structure of the Kaggle Black Box Learning Challenge
data set.\protect \\
The KNN with Euclidean metric (green continuous line) and with cosine metric
(blue dashed line) has been used to explore the structure of the data. The
plot shows the average accuracy and the standard error of KNN (over
five simulations). \label{fig:Plot-BB}}
\end{figure}

Figure \ref{fig:Plot-BB} confirms our intuition. For all the different
values of $k$ we considered, the cosine distance proved to be a better
metric to explain the structure of the data in the Kaggle Black Box
Learning Challenge. This provides an explanation why sparse filtering
proved so useful with the KBBLC data, when compared to other standard
unsupervised learning algorithms, especially those based on the Euclidean
metric. This result agrees with the fact that the Euclidean metric
is not a suitable metric for measuring distances among samples of 
digits represented in the pixel space \citep{Simard1998}.

\section{Discussion \label{sec:Discussion}}

Our theoretical and empirical analysis showed that the standard sparse filtering algorithm implemented with an absolute-value non-linearity preserves the data structure explained by the cosine neighborhoodness In our experiments, we have shown that, when the relevant structure of the data has a radial structure, then sparse filtering may be expected to perform significantly better than the standard Euclidean-based alternatives. 
Indeed, sparse filtering may be seen as an algorithm approximately transforming cosine distances in the original space into Euclidean distances in the representation space.  

It is normally assumed that the data points $\mathbf{X}^{(i)}$ are best explained as samples from a multivariate random variable $X=\left(X_{1},X_{2},\ldots,X_{O}\right)$, where each random variable $X_{j}$ describes a component $\mathbf{X}_{j}^{(i)}$. However, given the data points $\mathbf{X}^{(i)}$, it is possible to assume that the generating process is better described by a multivariate random variable $X^{'}=\left(X_{1}^{'},X_{2}^{'},\ldots,X_{O-1}^{'}\right)$, where each random variable describes an angular coordinate $\theta_{j}$ of $\mathbf{X}_{j}^{(i)}$. Sparse filtering tries to preserve the information about the $O-1$ angular coordinates $\theta_{i}$, discarding the information about the radial coordinate $\rho$. If $p(Y \vert X)$ is better explained in terms of radial coordinates, then sparse filtering is a very reasonable choice for unsupervised representation learning.\\

Our study allow us to conclude that our original thesis is correct: sparse filtering satisfies both the informativeness principle and the infomax principle. 
In particular, the informativeness principle is satisfied through the adoption of the proxy of sparsity, as shown in section \ref{sec:informativeness}. 
The infomax principle is satisfied through the preservation of a precise structure underlying the data, that is, the radial structure of the data. Mutual information between the original representations $\mathbf{X}^{(i)}$ and the learned representations $\mathbf{Z}^{(i)}$ is retained when the structure of the data is explained by the cosine metric, that is, in an ideal case, when all the information is carried by the angular coordinates of the data, as demonstrated in section \ref{sec:pres-cosine-neigh}. Indeed, the mutual information between the original and the learned representations can be formally expressed as: $I\left[X,Z\right] = H\left[X\right] - H\left[X \vert Z\right]$. Given that the entropy of the distribution of the data $p(X)$ is fixed, the only way to maximize the mutual information is by minimizing the conditional entropy $H\left[X \vert Z\right]$. Since the representation $\mathbf{Z}$ preserve all the information about the angular coordinates of the original data, the uncertainty about $\mathbf{X}$ given $\mathbf{Z}$ is minimized if the structure of the data has indeed a radial structure.\\

Following this reasoning, we suggested an interpretation of sparse filtering as an unsupervised soft clustering algorithm based on the cosine metric.
This perspective allowed us to contrast the results of sparse filtering with other standard algorithms for clustering based on the Euclidean metric and  conclude that sparse filtering does not provide a better processing of the data in absolute terms, but instead it offers an alternative interpretation of the data based on a different metric.\\

While in our experiments, we were aware a priori of the metric underlying synthetic data sets, in a real-world setting such knowledge may not be available and simple exploratory analysis of the data (using, for instance KNN) may be unsuitable. Sparse filtering, thanks to its scalability and its efficiency, could also be used to infer the data structure underlying the data. The usefulness of polar coordinates in several scientific fields and physical applications may suggest that interpreting data according to cosine distance could be a sensible choice.\\

Additionally, we proved that, in high dimensions, sparse filtering can still probabilistically preserve Euclidean distances. This is justified by the fact that, under the assumptions we made, the probability that unrelated points with high Euclidean distance will have the same angular coordinates $\theta_{i}$ can be bounded (Section \ref{sec:pres-euclidean}).\\

 Interestingly, our study of sparse filtering as an unsupervised learning algorithm shares a similar methodology with the very recent work by \citet{McNamara2016}. Re-casting our analysis in their framework we can demonstrate that, with high probability, sparse filtering working on radial data contributes to the reduction of the risk in standard supervised learners by showing that: (i) $P(X)$ has a structure explained by cosine neighborhoodness; (ii) $P(Y \vert X)$ share the same structure as $P(X)$; (iii) sparse filtering relies on the cosine distance; (iv) a supervised learner, such as SVM, can exploit the new Euclidean structure in the learned representations.

\section{Concluding Remarks\label{sec:Conclusion}}

In this paper, we have explained \emph{why} sparse filtering works
(by proving its property of preservation of cosine neighborhoodness)
and \emph{when} it should be expected to provide useful representations
(by considering the data structure of the samples).\\

Our theoretical analysis and simulations were not designed to show 
that sparse filtering is able to provide state-of-the-art performance
against other algorithms, but, instead, to show how the implicit assumptions
and constraints of sparse filtering make it better suited for certain
scenarios instead of others.  In particular, for the standard sparse filtering algorithm implemented using an absolute-value non-linearity we demonstrated that its success is tied to the data having a structure explained by cosine neighborhoodness.
Consistently with the no-free lunch theorem \citep{Wolpert1997}, we reached the conclusion that sparse filtering is not a better algorithm than other Euclidean-based clustering algorithms, but that there is a specific set of problems (in which $p(Y \vert X)$ is explained by the cosine metric) where the performance of sparse filtering is excellent, balanced by a set of problems (in which $p(Y \vert X)$ is explained by the Euclidean metric or other metrics) where its performance is less outstanding.
This led us to interpret the representation of sparse filtering as a ``view'' of the data according to the cosine metric alternative to the more standard Euclidean view. Combining these two different views could provide representations with more discriminative power.\\

At the foundation of our analysis lies the understanding that sparse
filtering \textit{must} preserve some information carried by the pdf $p(X)$. Despite sparse filtering
ignoring the problem of explicitly modeling the true pdf $p\left(X\right)$,
we showed that the algorithm is hard-coded with an implicit constraint that guarantees the preservation of some data structure. This is clearly a specific
conclusion about the particular algorithm of sparse filtering, but
we can expect that this principle will be applicable to the whole
class of feature distribution learning algorithms. We might expect
that any feature distribution learning algorithm, in order to be successful,
must take into account, through constraints or priors, the problem
of preserving the mutual information between the original representations
$\mathbf{X}^{(i)}$ and the learned representations $\mathbf{Z}^{(i)}$.

Being aware of this requirement can give a precious contribution in the future research and design of new feature distribution learning algorithms; for instance, it may prevent us from considering in sparse filtering alternative non-linearities that do not preserve any interesting structure (such as, the ReLU function) or it may help us to avoid solutions that, being unable to preserve any structure of the data, are bound to produce unsatisfactory representations. Ongoing research is focused on discovering which structures may be preserved by alternative version of sparse filtering, with a particular focus on periodic structures that may be learned using trigonometric functions.\\

A deeper theoretical understanding of the dynamics of sparse filtering may be developed in connection with manifold learning and information geometry \citep{Amari2016}. The property of preservation of structure that we uncovered in this paper may be more formally explained in the framework of differential geometry, by modeling the data samples as a points on a Riemannian manifold. Relevant data structures (that we presented in terms of Euclidean or cosine distance) may then be described in terms of Riemannian metric tensors, and preservation properties may be studied in terms of preservation of these tensors.\\ 

Another promising avenue in our research is the extension
of sparse filtering to semi-supervised learning. Indeed, the paradigm
of feature distribution learning seems perfectly suited for the scenario
in which we are provided with few labeled samples and many unlabeled
samples: we 
may exploit the information carried by the labeled samples to better shape
the feature distribution $p(Z)$, without addressing the problem of
estimating the true pdf $p\left(X^{*}\right)$; at the same time,
the constraint of sparsity would help us to not overfit, and the constraint
of structure preservation would help us to preserve the information
conveyed by $p(X)$. Furthermore, assuming some regularity in the
original representation space, we hypothesize that we could use the
information in the labeled samples to address the problem of covariate
shift \citep{Sugiyama2012} in a semi-supervised learning scenario.

\appendix

\section{Proofs}

\subsection{Proposition \ref{non-preservation-euclidean} \label{app:non-preservation-euclidean}}

\textbf{Proposition 1.}
\emph{	
	Let $\left\{ \mathbf{X}^{(i)}\in\mathbb{R}^{O}\right\} _{i=1}^{N}$
	be a set of points in the original space $\mathbb{R}^{O}$. Then,
	the transformations from A1 to A4 do not preserve the structure of
	the data described by the Euclidean metric.}

\textbf{Proof.} We prove this proposition by counterexample.

Let us consider the case in which $\mathbf{X}^{(1)}$ is a vector such that 
$\mathbf{X}^{(1)}_j = \frac{1}{\sqrt{2}}$, $\forall j$,  $1 \leq j \leq O$, $\mathbf{X}^{(2)}$ is another vector such that $\mathbf{X}^{(2)}=-\mathbf{X}^{(1)}$, $L = O$,
and $\mathbf{W}=\mathbf{I}$, where $\mathbf{I}$ is the identity
matrix.

The Euclidean distance between the vectors $\mathbf{X}^{(1)}$ and $\mathbf{X}^{(2)}$ is:
\[
D_{E}\left(\mathbf{X}^{(1)},\mathbf{X}^{(2)}\right)=\sqrt{\sum_{j=1}^{O}\left(\frac{1}{\sqrt{2}}+\frac{1}{\sqrt{2}}\right)^{2}}=\sqrt{2L}.
\]
Let us now apply the transformation $f_{A1:A4}$ to the vectors $\mathbf{X}^{(1)}$ and $\mathbf{X}^{(2)}$:
\[
\begin{array}{ccc}
f_{A1}\left(\mathbf{X}^{(1)}\right)=\mathbf{I}\mathbf{X}^{(1)}=\mathbf{X}^{(1)} &  & f_{A1}\left(\mathbf{X}^{(2)}\right)=\mathbf{I}\mathbf{X}^{(2)}=\mathbf{X}^{(2)}\\

f_{A2}\left(\mathbf{X}^{(1)}\right)=\left|\mathbf{X}^{(1)}\right|=\mathbf{X}^{(1)} &  & f_{A2}\left(\mathbf{X}^{(2)}\right)=\left|\mathbf{X}^{(2)}\right|=\mathbf{\mathbf{X}}^{(1)}\\

f_{A3}\left(\mathbf{X}^{(1)}\right)=\left[\frac{\mathbf{X}_{j}^{(1)}\sqrt{2}}{\sqrt{N}}\right]=\left[\frac{1}{\sqrt{N}}\right] &  & f_{A3}\left(\mathbf{X}^{(1)}\right)=\left[\frac{\mathbf{X}_{j}^{(1)}\sqrt{2}}{\sqrt{N}}\right]=\left[\frac{1}{\sqrt{N}}\right]\\

f_{A4}\left(\left[\frac{1}{\sqrt{N}}\right]\right)=\left[\frac{\mathbf{X}_{j}^{(1)}\sqrt{N}}{\sqrt{L}}\right]=\left[\frac{1}{\sqrt{L}}\right]=\mathbf{Z}^{(1)} &  & f_{A4}\left(\left[\frac{1}{\sqrt{N}}\right]\right)=\left[\frac{\mathbf{X}_{j}^{(1)}\sqrt{N}}{\sqrt{L}}\right]=\left[\frac{1}{\sqrt{L}}\right]=\mathbf{Z}^{(1)}.
\end{array}
\]
Thus, $f_{A1:A4}\left(\mathbf{X}^{(1)}\right)=\mathbf{Z}^{(1)}$ and $f_{A1:A4}\left(\mathbf{X}^{(2)}\right)=\mathbf{Z}^{(1)}$.
Now, the Euclidean distance between the vectors $f_{A1:A4}\left(\mathbf{X}^{(1)}\right)$
and $f_{A1:A4}\left(\mathbf{X}^{(2)}\right)$ is:
\[
D_{E}\left(\mathbf{Z}^{(1)},\mathbf{Z}^{(1)}\right)=0.
\]
Therefore the transformations from A1 to A4 do not preserve the structure
of the data described by the Euclidean metric. $\blacksquare$

\subsection{Lemma \ref{lem1}} \label{app:lem1}
\textbf{Lemma 1.}
\emph{Let us consider $\mathbf{u},\mathbf{v}\in\mathbb{R}^{O}$, two generic collinear vectors, and let $f:\mathbb{\mathbb{R}}^{O}\rightarrow\mathbb{R}^{L}$	be a linear transformation defined as $f(\mathbf{u})=\mathbf{Wu}$, where $\mathbf{W}$ is the matrix associated with the linear transformation. Then, $f\left(\mathbf{u}\right),f(\mathbf{v})\in\mathbb{R}^{L}$ are also collinear.}

\textbf{Proof.} 
Let us consider the two collinear vectors $\mathbf{u}$ and $\mathbf{v}$.
By definition, collinearity means that there exists $k\in\mathbb{R}$,
$k\neq0$, such that $\mathbf{v}=k\mathbf{u}$.
Let us now consider the linear transformation $f$ encoded by matrix
$\mathbf{W}$ and let us apply it to the vector $\mathbf{v}$:
\[
f(\mathbf{v})=\mathbf{W}\mathbf{v}=\mathbf{W}\left(k\mathbf{u}\right)=k\left(\mathbf{Wu}\right)=k\cdot f(\mathbf{u}).
\]
Therefore, collinearity is preserved.  $\blacksquare$

\subsection{Lemma \ref{lem2}} \label{app:lem2}
\textbf{Lemma 2.}
\emph{Let us consider $\mathbf{u},\mathbf{v}\in\mathbb{R}^{L}$, two generic collinear vectors, and let $f:\mathbb{\mathbb{R}}^{L}\rightarrow\mathbb{R}^{L}$ be the element-wise absolute-value function $f(\mathbf{u})=\left|\mathbf{u}\right|=\left[\left|u_{j}\right|\right]$. Then $f\left(\mathbf{u}\right),f(\mathbf{v})\in\mathbb{R}^{L}$ 	are also collinear.}

\textbf{Proof.} 
Let us consider the two collinear vectors $\mathbf{u}$ and $\mathbf{v}$.
By definition, collinearity means that there exists $k\in\mathbb{R}$,
$k\neq0$, such that $\mathbf{v}=k\mathbf{u}$.
Let us now consider the element-wise absolute-value function $f$
and let us apply it to the vector $\mathbf{v}$:
\[
f(\mathbf{v})=\left|\mathbf{v}\right|=\left|k\mathbf{u}\right|=\left|k\right|\cdot\left|\mathbf{u}\right|=\left|k\right|\cdot f(\mathbf{u}).
\]
Therefore, collinearity is preserved. $\blacksquare$

\subsection{Lemma \ref{lem3}} \label{app:lem3}
\textbf{Lemma 3.}
\emph{Let us consider $\mathbf{u},\mathbf{v}\in\mathbb{R}^{L}$, two collinear vectors whose components are all strictly positive\footnote{Notice that we can safely make the assumption of strict positivity in sparse filtering since $\mathbf{u}$ and $\mathbf{v}$ are the output of an absolute-value function implemented as $f(x)=\sqrt{x^{2}+\epsilon}$.}, and let $f:\mathbb{\mathbb{R}}^{L}\rightarrow\mathbb{R}^{L}$ be the $\ell_{2}$-normalization along the features. Then $f\left(\mathbf{u}\right),f(\mathbf{v})\in\mathbb{R}^{L}$ are also collinear.}

\textbf{Proof.} Let us consider the two collinear vectors $\mathbf{u}$ and $\mathbf{v}$.
By definition, collinearity means that there exists $k\in\mathbb{R}$,
$k\neq0$, such that $\mathbf{v}=k\mathbf{u}$.
Let us now consider the function of normalization along the features
$f(\mathbf{u})=\left[\frac{\mathbf{u}_{j}}{\sqrt{\sum_{\mathbf{w}=\{\mathbf{u},\mathbf{v}...\}}\mathbf{w}_{j}^{2}}}\right]$.
Normalizing along the features means dividing each component $\mathbf{u}_{j}$
by a constant $c_{j}$ equal to the $\ell_{2}$-norm of the component
$j$ across all the available vectors $\mathbf{u},\mathbf{v},...$,
that is $f(\mathbf{u})=\left[\frac{\mathbf{u}_{j}}{c_{j}}\right]=\mathbf{c}\circ\mathbf{u}$,
where $\mathbf{c}$ is the vector containing all the constants $c_{j}$
and $\circ$ is the element-wise product. Let us now apply the normalization
along the features to the vector $\mathbf{v}$:
\[
f(\mathbf{v})=\mathbf{c}\circ\mathbf{v}=\mathbf{c}\circ\left(k\mathbf{u}\right)=k\cdot\left(\mathbf{c}\circ\mathbf{u}\right)=k\cdot f(\mathbf{u}).
\]
Therefore, collinearity is preserved. $\blacksquare$

\subsection{Lemma \ref{lem4}} \label{app:lem4}
\textbf{Lemma 4.}
\emph{Let us consider $\mathbf{u}\in\mathbb{R}^{L}$, a vector whose components are all strictly positive \footnote{Notice that we can safely make the assumption of strict positivity in sparse filtering since $\mathbf{u}$ is the output of the normalization along the feature which preserves the positivity.}, and let  $f:\mathbb{\mathbb{R}}^{L}\rightarrow\mathbb{R}^{L}$ be the $\ell_{2}$-normalization along the samples. Then $f\left(\mathbf{u}\right)\in\mathbb{R}^{L}$ have the same angular coordinates as $\mathbf{u}$. }

\textbf{Proof.} 
Let us consider the function of normalization along the features $f(\mathbf{u})=\left[\frac{\mathbf{u}_{j}}{\sqrt{\sum_{j}\mathbf{u}_{j}^{2}}}\right]$.
Normalizing along the samples means dividing each component $\mathbf{u}_{j}$
by the $\ell_{2}$-norm of the same vector $\mathbf{u}$, that is
$f(\mathbf{u})=\left[\frac{\mathbf{u}_{j}}{\ell_{2}(\mathbf{u})}\right]=\frac{1}{\ell_{2}(\mathbf{u})}\cdot\mathbf{u}$.
Multiplying all the components of the same vector $\mathbf{u}$ by
the constant $k=\frac{1}{\ell_{2}(\mathbf{u})}$ leaves the angular
coordinates unaltered. 
Therefore, the angular coordinates are preserved. $\blacksquare$

\subsection{Theorem \ref{preservation-cosine-neigh} \label{app:preservation-cosine-neigh}}

\textbf{Theorem 4.}
\emph{ 
	Let $\mathbf{X}^{(1)},\mathbf{X}^{(2)}\in\mathbb{R}^{O}$
	be two original data samples and let $\mathbf{Z}^{(1)},\mathbf{Z}^{(2)}\in\mathbb{R}^{L}$
	be their representations computed by sparse filtering. If the cosine
	distance between the original samples is arbitrarily small $D_{C}\left[\mathbf{X}^{(1)},\mathbf{X}^{(2)}\right]<\delta$,
	for $\delta>0$, then the Euclidean distance between the computed
	representations is arbitrarily small $D_{E}\left[\mathbf{Z}^{(1)},\mathbf{Z}^{(2)}\right]<\epsilon$,
	for $\epsilon>0$, and $\epsilon=L \cdot \left(\frac{k+\left|\sqrt{2\delta-\delta^{2}}\right|}{\ell_{2}\left(\tilde{\mathbf{F}}^{(2)}\right)}-\frac{1}{\ell_{2}\left(\tilde{\mathbf{F}}^{(1)}\right)}\right)$,
	where $k$ is a constant accounting for partial collinearity and $\ell_{2}\left(\tilde{\mathbf{F}}^{(i)}\right)$
	is the $\ell_{2}$-norm of the representations computed by sparse
	filtering after step A3.
}

\textbf{Proof.} In order to prove this theorem we adopt the following strategy:
we compute the representations at each step of the computation (before
sparse filtering, after steps A1 and A2, after step A3 and after step
A4) and we upper bound the displacement accounting for the Euclidean
distance between the representations.

Recall that given two generic points $\mathbf{X}^{(1)}$ and $\mathbf{X}^{(2)}$,
we can express $\mathbf{X}^{(2)}$ as a function of $\mathbf{X}^{(1)}$
plus a \emph{displacement} vector $\bar{\mathbf{X}}$:
\begin{equation}
\mathbf{X}^{(2)}=\mathbf{X}^{(1)}+\bar{\mathbf{X}},\label{eq:displacement0}
\end{equation}
so that we can easily account for the Euclidean distance between $\mathbf{X}^{(1)}$
and $\mathbf{X}^{(2)}$ just as a function of the displacement vector
$\bar{\mathbf{X}}$:
\[
D_{E}\left[\mathbf{X}^{(1)},\mathbf{X}^{(2)}\right]=\ell_{2}\left(\bar{\mathbf{X}}\right).
\]

\textbf{(Before sparse filtering.)} Let us now consider two points
$\mathbf{X}^{(1)}$ and $\mathbf{X}^{(2)}$ which are almost collinear
with an arbitrary small cosine distance \emph{$D_{C}\left[\mathbf{X}^{(1)},\mathbf{X}^{(2)}\right]<\delta$.
}We can then express $\mathbf{X}^{(2)}$ as a point collinear with $\mathbf{X}^{(1)}$
to which a \emph{bias} vector $\mathbf{B}$ is added:
\[
\mathbf{X}^{(2)}=k\mathbf{X}^{(1)}+\mathbf{B},
\]
where $k\in\mathbb{R}$ is a constant that preserves collinearity.
With no loss of generality, we will assume $k>1$; we exclude values
of $k$ smaller than zero which would generate a reflection (reflections
are not relevant for the following treatment as they induce a cosine
distance far greater than $\delta$) and we ignore values of $k$
between zero and one (in such a case, our proof will hold once
we swap $\mathbf{X}^{(1)}$ and $\mathbf{X}^{(2)}$). The bias vector
$\mathbf{B}$ accounts for a relative displacement between the perfectly
collinear sample $k\mathbf{X}^{(1)}$ and the almost collinear sample
$k\mathbf{X}^{(1)}+\mathbf{B}$.

With reference to Equation \ref{eq:displacement0}, the displacement
vector $\bar{\mathbf{X}}$ is:
\begin{equation}
\bar{\mathbf{X}}=(k-1)\mathbf{X}^{(1)}+\mathbf{B},\label{eq:displacement1}
\end{equation}
from which follows that:
\[
D_{E}\left[\mathbf{X}^{(1)},\mathbf{X}^{(2)}\right]=\ell_{2}\left(\bar{\mathbf{X}}\right)=\ell_{2}\left((k-1)\mathbf{X}^{(1)}+\mathbf{B}\right).
\]

\textbf{(Before sparse filtering - Upper bound)} To upper bound $D_{E}\left[\mathbf{X}^{(1)},\mathbf{X}^{(2)}\right]$,
we can evaluate the maximum value that $\ell_{2}\left(\bar{\mathbf{X}}\right)$
can reach, consistent with the constraint of a bounded cosine distance
$D_{C}\left[\mathbf{X}^{(1)},\mathbf{X}^{(2)}\right]$. Formally,
we can set up the optimization problem:
\[
\max_{\bar{\mathbf{X}}\in\mathbb{R}^{O}}\ell_{2}\left(\bar{\mathbf{X}}\right)
\]
under the constraint:
\[
D_{C}\left[\mathbf{X}^{(1)},\mathbf{X}^{(2)}\right]<\delta.
\]
The maximization can be rewritten as: 
\begin{eqnarray*}
	\max_{\bar{\mathbf{X}}\in\mathbb{R}^{O}}\ell_{2}\left(\bar{\mathbf{X}}\right) & = & \max_{\bar{\mathbf{X}}_{j}\in\mathbb{R}}\sqrt{\sum_{j=1}^{O}\bar{\mathbf{X}}_{j}^{2}}\\
	& = & \max_{\mathbf{B}_{j}\in\mathbb{R}}\sqrt{\sum_{j=1}^{O}\left((k-1)\mathbf{X}_{j}^{(1)}+\mathbf{B}_{j}\right)^{2}}\\
	& = & \max_{\mathbf{B}_{j}\in\mathbb{R}}\sqrt{\sum_{j=1}^{O}\left(\mathbf{B}_{j}\right)^{2}}\\
	& = & \max_{\mathbf{B}_{j}\in\mathbb{R}}\mathbf{B}_{j},
\end{eqnarray*}
assuming: (i) that $\mathbf{X}^{(1)}$ and $k$ are given, and (ii)
that $\mathbf{X}_{j}^{(1)}$ and $\mathbf{B}_{j}$ are both positive
(as this constitute the worst case that needs to be considered in
the analysis of the upper bound). An upper bound on the displacement
$\bar{\mathbf{X}}$ can be then computed from the solution to the
individual constrained optimization problems for each component $\mathbf{B}_{j}$:
\[
\max_{\mathbf{B}_{j}\in\mathbb{R}}\mathbf{B}_{j}
\]
under the constraint:
\begin{eqnarray*}
	\delta & > & D_{C}\left[\mathbf{X}^{(1)},\mathbf{X}^{(2)}\right]\\
	& = & D_{C}\left[\mathbf{X}^{(1)},k\mathbf{X}^{(1)}+\mathbf{B}\right].
\end{eqnarray*}
By construction, we know that $D_{C}\left[\mathbf{X}^{(1)},k\mathbf{X}^{(1)}\right]=0$.
Therefore the entire cosine distance must be accounted by the bias
vector $\mathbf{B}$. Trigonometrically, from the cosine distance
$\delta$ we can recover the angle opposite to a cathetus corresponding
to the radius of an hypersphere centered on $k\mathbf{X}^{(1)}$ and
bounding the module of $\mathbf{B}$. Let $\theta$ be the underlying
angle between $\mathbf{X}^{(1)}$ and $\mathbf{X}^{(2)}$:
\begin{eqnarray*}
	\delta & = & 1-\cos\theta\\
	\theta & = & \arccos(1-\delta).
\end{eqnarray*}
The radius of the hypersphere centered on $k\mathbf{X}^{(1)}$ inducing
at most a cosine distance $\delta$ is:
\begin{eqnarray*}
	\mathbf{B}_{j} & \leq & \mathbf{X}_{j}^{(1)}\sin\arccos(1-\delta)\\
	& = & \mathbf{X}_{j}^{(1)}\sqrt{1-(1-\delta)^{2}}\\
	& = & \mathbf{X}_{j}^{(1)}\sqrt{2\delta-\delta^{2}}.
\end{eqnarray*}
Substituting in Equation \ref{eq:displacement1}, the displacement
on each component can the be upper bounded as:
\begin{eqnarray*}
	\bar{\mathbf{X}}_{j} & = & (k-1)\mathbf{X}_{j}^{(1)}+\mathbf{B}_{j}\\
	& \leq & (k-1)\mathbf{X}_{j}^{(1)}+\mathbf{X}_{j}^{(1)}\sqrt{2\delta-\delta^{2}}\\
	& = & \mathbf{X}_{j}^{(1)}\left(k-1+\sqrt{2\delta-\delta^{2}}\right).
\end{eqnarray*}
This upper bound depends on the original cosine distance $\delta$,
but, more significantly on the module of $\mathbf{X}^{(1)}$ and the
stretching constant $k$. Indeed, the Euclidean distance along each component
is given by the stretch ($\mathbf{X}_{j}^{(1)}\left(k-1\right)$)
plus a small distance due to the angle ($\mathbf{X}_{j}^{(1)}\sqrt{2\delta-\delta^{2}}$).

\textbf{(Steps A1 and A2)} Let us now apply the linear projection
and the absolute-value function defined in transformation A1 and A2:
\begin{eqnarray*}
	\mathbf{F}^{(1)}=f_{A1:A2}\left(\mathbf{X}^{(1)}\right) & = & \left|\mathbf{W}\mathbf{X}^{(1)}\right|\\
	\mathbf{F}^{(2)}=f_{A1:A2}\left(\mathbf{X}^{(2)}\right) & = & \left|\mathbf{W}\left(k\mathbf{X}^{(1)}+\mathbf{B}\right)\right|=k\mathbf{F}^{(1)}\pm\left|\mathbf{W}\mathbf{B}\right|.
\end{eqnarray*}
Component-wise we have:
\begin{eqnarray*}
	\mathbf{F}_{l}^{(1)} & = & \left|\sum_{j=1}^{O}\mathbf{W}_{l}^{(j)}\mathbf{X}_{j}^{(1)}\right|\\
	\mathbf{F}_{l}^{(2)} & = & k\mathbf{F}_{l}^{(1)}+\left|\mathbf{W}\mathbf{B}\right|_{l}=k\left|\sum_{j=1}^{O}\mathbf{W}_{l}^{(j)}\mathbf{X}_{j}^{(1)}\right|\pm\left|\sum_{j=1}^{O}\mathbf{W}_{l}^{(j)}\mathbf{B}_{j}\right|.
\end{eqnarray*}
The new displacement and the new Euclidean distance are:
\begin{equation}
\bar{\mathbf{F}}_{l}=\left(k-1\right)\mathbf{F}_{l}^{(1)}\pm\left|\mathbf{W}\mathbf{B}\right|_{l},\label{eq:displacement2}
\end{equation}
\[
D_{E}\left[\mathbf{F}^{(1)},\mathbf{F}^{(2)}\right]=\ell_{2}\left(\bar{\mathbf{F}}\right)=\sqrt{\sum_{l=1}^{L}\left(\left(k-1\right)\mathbf{F}_{l}^{(1)}\pm\left|\mathbf{W}\mathbf{B}\right|_{l}\right)^{2}}.
\]

\textbf{(Steps A1 and A2 - Upper bound)} The upper bound of each component
of the new bias vector follows immediately:
\begin{eqnarray*}
	\left|\mathbf{W}\mathbf{B}\right|_{l} & = & \left|\sum_{j=1}^{O}\mathbf{W}_{l}^{(j)}\mathbf{B}_{j}\right|\\
	& \leq & \left(\left|\sum_{j=1}^{O}\mathbf{W}_{l}^{(j)}\mathbf{X}_{j}^{(1)}\sqrt{2\delta-\delta^{2}}\right|\right)\\
	& = & \left|\sqrt{2\delta-\delta^{2}}\right|\left|\sum_{j=1}^{O}\mathbf{W}_{l}^{(j)}\mathbf{X}_{j}^{(1)}\right|,
\end{eqnarray*}
and then the upper bound on each component of the displacement in
Equation \ref{eq:displacement2} is:
\begin{eqnarray*}
	\bar{\mathbf{F}}_{l} & \leq & \left(k-1\right)\mathbf{F}_{l}^{(1)}+\left|\sqrt{2\delta-\delta^{2}}\right|\left|\sum_{j=1}^{O}\mathbf{W}_{l}^{(j)}\mathbf{X}_{j}^{(1)}\right|\\
	& = & \left(k-1\right)\left|\sum_{j=1}^{O}\mathbf{W}_{l}^{(j)}\mathbf{X}_{j}^{(1)}\right|+\left|\sqrt{2\delta-\delta^{2}}\right|\left|\sum_{j=1}^{O}\mathbf{W}_{l}^{(j)}\mathbf{X}_{j}^{(1)}\right|\\
	& = & \left(k-1+\left|\sqrt{2\delta-\delta^{2}}\right|\right)\left|\sum_{j=1}^{O}\mathbf{W}^{(l)}\mathbf{X}_{j}^{(1)}\right|.
\end{eqnarray*}

\textbf{(Step A3)} Let us now apply the normalization along the rows
defined in transformation A3:
\begin{eqnarray*}
	\tilde{\mathbf{F}}_{l}^{(1)}=f_{A3}\left(\mathbf{F}_{l}^{(1)}\right) & = & \frac{\mathbf{F}_{l}^{(1)}}{\sqrt{\sum_{i}^{N}\left(\mathbf{F}_{l}^{(i)}\right)^{2}}}\\
	\tilde{\mathbf{F}}_{l}^{(2)}=f_{A3}\left(\mathbf{F}_{l}^{(2)}\right) & = & \frac{k\mathbf{F}_{l}^{(1)}+\left|\mathbf{W}\mathbf{B}\right|_{l}}{\sqrt{\sum_{i}^{N}\left(\mathbf{F}_{l}^{(i)}\right)^{2}}}=k\tilde{\mathbf{F}}_{l}^{(1)}+\frac{\left|\mathbf{W}\mathbf{B}\right|_{l}}{\sqrt{\sum_{i}^{N}\left(\mathbf{F}_{l}^{(i)}\right)^{2}}}.
\end{eqnarray*}
Notice that the denominator is given by a feature-dependent sum across
$N$ samples; for simplicity, we will take this value to be a constant
$\left\{ c_{l}\right\} _{l=1}^{L}$, $c\in\mathbb{R}$:
\begin{eqnarray*}
	\tilde{\mathbf{F}}_{l}^{(1)} & = & \frac{\mathbf{F}_{l}^{(1)}}{c_{l}}\\
	\tilde{\mathbf{F}}_{l}^{(2)} & = & k\tilde{\mathbf{F}}_{l}^{(1)}+\frac{\left|\mathbf{W}\mathbf{B}\right|_{l}}{c_{l}}.
\end{eqnarray*}
The new displacement and the new Euclidean distance are:
\begin{equation}
\bar{\mathbf{\tilde{F}}}_{l}=\left(k-1\right)\tilde{\mathbf{F}}_{l}^{(1)}+\frac{\left|\mathbf{W}\mathbf{B}\right|_{l}}{c_{l}},\label{eq:displacement3}
\end{equation}
\[
D_{E}\left[\tilde{\mathbf{F}}^{(1)},\tilde{\mathbf{F}}^{(2)}\right]=\ell_{2}\left(\bar{\mathbf{\tilde{F}}}\right)=\sqrt{\sum_{l=1}^{L}\left(\left(k-1\right)\tilde{\mathbf{F}}_{l}^{(1)}+\frac{\left|\mathbf{W}\mathbf{B}\right|_{l}}{c_{l}}\right)^{2}}.
\]

\textbf{(Step A3 - Upper bound)} The upper bound of each component
of the new bias vector follows immediately:
\begin{eqnarray*}
	\frac{\left|\mathbf{W}\mathbf{B}\right|_{l}}{c_{l}} & \leq & \frac{\left|\sqrt{2\delta-\delta^{2}}\right|\left|\sum_{j=1}^{O}\mathbf{W}_{l}^{(j)}\mathbf{X}_{j}^{(1)}\right|}{c_{l}}
\end{eqnarray*}
and then the upper bound on each component of the displacement in
Equation \ref{eq:displacement3}:
\begin{eqnarray*}
	\bar{\mathbf{\tilde{F}}}_{l} & \leq & \left(k-1\right)\tilde{\mathbf{F}}_{l}^{(1)}+\frac{\left|\sqrt{2\delta-\delta^{2}}\right|\left|\sum_{j=1}^{O}\mathbf{W}_{l}^{(j)}\mathbf{X}_{j}^{(1)}\right|}{c_{l}}\\
	& = & \left(k-1\right)\frac{\mathbf{F}_{l}^{(1)}}{c_{l}}+\frac{\left|\sqrt{2\delta-\delta^{2}}\right|\left|\sum_{j=1}^{O}\mathbf{W}_{l}^{(j)}\mathbf{X}_{j}^{(1)}\right|}{c_{l}}\\
	& = & \left(k-1\right)\frac{\left|\sum_{j=1}^{O}\mathbf{W}_{l}^{(j)}\mathbf{X}_{j}^{(1)}\right|}{c_{l}}+\frac{\left|\sqrt{2\delta-\delta^{2}}\right|\left|\sum_{j=1}^{O}\mathbf{W}_{l}^{(j)}\mathbf{X}_{j}^{(1)}\right|}{c_{l}}\\
	& = & \frac{k-1+\left|\sqrt{2\delta-\delta^{2}}\right|}{c_{l}}\left|\sum_{j=1}^{O}\mathbf{W}_{l}^{(j)}\mathbf{X}_{j}^{(1)}\right|\\
	& = & \frac{1}{c_{l}}\bar{\mathbf{F}}_{l}.
\end{eqnarray*}
Not surprisingly, after transformation A3, the Euclidean distance
$D_{E}\left[\tilde{\mathbf{F}}^{(1)},\tilde{\mathbf{F}}^{(2)}\right]$
is just rescaled since each component of the displacement $\bar{\mathbf{F}}_{l}$
is reduced by a factor $\frac{1}{c_{l}}=\frac{1}{\sqrt{\sum_{i}^{N}\left(\mathbf{F}_{l}^{(i)}\right)^{2}}}$
.

\textbf{(Step A4)} Finally, let us apply the normalization along the
samples defined in transformation A4:
\begin{eqnarray*}
	\mathbf{Z}_{l}^{(1)} & = & f_{A4}\left(\tilde{\mathbf{F}}_{l}^{(1)}\right)=\frac{\tilde{\mathbf{F}}_{l}^{(1)}}{\ell_{2}\left(\tilde{\mathbf{F}}^{(1)}\right)}=\frac{\frac{\mathbf{F}_{l}^{(1)}}{c_{l}}}{\ell_{2}\left(\tilde{\mathbf{F}}^{(1)}\right)}\\
	\mathbf{Z}_{l}^{(2)} & = & f_{A4}\left(\tilde{\mathbf{F}}_{l}^{(2)}\right)=\frac{\tilde{\mathbf{F}}_{l}^{(2)}}{\ell_{2}\left(\tilde{\mathbf{F}}^{(2)}\right)}=\frac{k\tilde{\mathbf{F}}_{l}^{(1)}+\frac{\left|\mathbf{W}\mathbf{B}\right|_{l}}{c_{l}}}{\ell_{2}\left(\tilde{\mathbf{F}}^{(2)}\right)}=\frac{k\frac{\mathbf{F}_{l}^{(1)}}{c_{l}}}{\ell_{2}\left(\tilde{\mathbf{F}}^{(2)}\right)}+\frac{\frac{\left|\mathbf{W}\mathbf{B}\right|_{l}}{c_{l}}}{\ell_{2}\left(\tilde{\mathbf{F}}^{(2)}\right)}.
\end{eqnarray*}
Let us now consider the first term of $\mathbf{Z}_{l}^{(2)}$ and
let us multiply it by $\frac{\ell_{2}\left(\tilde{\mathbf{F}}^{(1)}\right)}{\ell_{2}\left(\tilde{\mathbf{F}}^{(1)}\right)}$:
\[
\mathbf{Z}_{l}^{(2)}=\frac{k\frac{\mathbf{F}_{l}^{(1)}}{c_{l}}}{\ell_{2}\left(\tilde{\mathbf{F}}^{(2)}\right)}\frac{\ell_{2}\left(\tilde{\mathbf{F}}^{(1)}\right)}{\ell_{2}\left(\tilde{\mathbf{F}}^{(1)}\right)}+\frac{\frac{\left|\mathbf{W}\mathbf{B}\right|_{l}}{c_{l}}}{\ell_{2}\left(\tilde{\mathbf{F}}^{(2)}\right)}=k\mathbf{Z}_{l}^{(1)}\frac{\ell_{2}\left(\tilde{\mathbf{F}}^{(1)}\right)}{\ell_{2}\left(\tilde{\mathbf{F}}^{(2)}\right)}+\frac{\frac{\left|\mathbf{W}\mathbf{B}\right|_{l}}{c_{l}}}{\ell_{2}\left(\tilde{\mathbf{F}}^{(2)}\right)}.
\]
The new displacement and the new Euclidean distance are:
\begin{equation}
\bar{\mathbf{Z}}_{l}=\left(k\frac{\ell_{2}\left(\tilde{\mathbf{F}}^{(1)}\right)}{\ell_{2}\left(\tilde{\mathbf{F}}^{(2)}\right)}-1\right)\mathbf{Z}_{l}^{(1)}+\frac{\left|\mathbf{W}\mathbf{B}\right|_{l}}{c_{l}\ell_{2}\left(\tilde{\mathbf{F}}^{(2)}\right)},\label{eq:displacement4}
\end{equation}
\begin{equation}
D_{E}\left[\mathbf{Z}^{(1)},\mathbf{Z}^{(2)}\right]=\ell_{2}\left(\bar{\mathbf{Z}}\right)=\sqrt{\sum_{l=1}^{L}\left(\left(k\frac{\ell_{2}\left(\tilde{\mathbf{F}}^{(1)}\right)}{\ell_{2}\left(\tilde{\mathbf{F}}^{(2)}\right)}-1\right)\mathbf{Z}^{(1)}+\frac{\left|\mathbf{W}\mathbf{B}\right|_{l}}{c_{l}\ell_{2}\left(\tilde{\mathbf{F}}^{(2)}\right)}\right)^{2}}.\label{eq:distance}
\end{equation}
For consistency, notice that if $\mathbf{X}^{(1)}$ and $\mathbf{X}^{(2)}$
were to be collinear, then $\ell_{2}\left(\tilde{\mathbf{F}}^{(2)}\right)=k\ell_{2}\left(\tilde{\mathbf{F}}^{(1)}\right)$,
and, by construction, $\mathbf{B}=0$; therefore, in case of collinearity,
$D_{E}\left[\mathbf{Z}^{(1)},\mathbf{Z}^{(2)}\right]$ computed in
Equation \ref{eq:distance} would be zero, thus agreeing with Theorem
\ref{thm-homo-representation-collinear-point}.

\textbf{(Step A4 - Upper bound)} Now, the upper bound of each component
of the bias vector can be immediately upper bounded:
\begin{eqnarray*}
	\frac{\left|\mathbf{W}\mathbf{B}\right|_{l}}{c_{l}\ell_{2}\left(\tilde{\mathbf{F}}^{(2)}\right)} & \leq & \frac{\left|\sqrt{2\delta-\delta^{2}}\right|\left|\sum_{j=1}^{O}\mathbf{W}_{l}^{(j)}\mathbf{X}_{j}^{(1)}\right|}{c_{l}\ell_{2}\left(\tilde{\mathbf{F}}^{(2)}\right)},
\end{eqnarray*}
and then the upper bound on each component of the displacement:
\begin{eqnarray*}
	\bar{\mathbf{Z}}_{l} & \leq & \left(k\frac{\ell_{2}\left(\tilde{\mathbf{F}}^{(1)}\right)}{\ell_{2}\left(\tilde{\mathbf{F}}^{(2)}\right)}-1\right)\mathbf{Z}_{l}^{(1)}+\frac{\left|\sqrt{2\delta-\delta^{2}}\right|\left|\sum_{j=1}^{O}\mathbf{W}_{l}^{(j)}\mathbf{X}_{j}^{(1)}\right|}{c_{l}\ell_{2}\left(\tilde{\mathbf{F}}^{(2)}\right)}\\
	& = & \left(k\frac{\ell_{2}\left(\tilde{\mathbf{F}}^{(1)}\right)}{\ell_{2}\left(\tilde{\mathbf{F}}^{(2)}\right)}-1\right)\frac{\mathbf{F}_{l}^{(1)}}{c_{l}\ell_{2}\left(\tilde{\mathbf{F}}^{(1)}\right)}+\frac{\left|\sqrt{2\delta-\delta^{2}}\right|\left|\sum_{j=1}^{O}\mathbf{W}_{l}^{(j)}\mathbf{X}_{j}^{(1)}\right|}{c_{l}\ell_{2}\left(\tilde{\mathbf{F}}^{(2)}\right)}\\
	& = & \left(k\frac{\ell_{2}\left(\tilde{\mathbf{F}}^{(1)}\right)}{\ell_{2}\left(\tilde{\mathbf{F}}^{(2)}\right)}-1\right)\frac{\left|\sum_{j=1}^{O}\mathbf{W}_{l}^{(j)}\mathbf{X}_{j}^{(1)}\right|}{c_{l}\ell_{2}\left(\tilde{\mathbf{F}}^{(1)}\right)}+\frac{\left|\sqrt{2\delta-\delta^{2}}\right|\left|\sum_{j=1}^{O}\mathbf{W}_{l}^{(j)}\mathbf{X}_{j}^{(1)}\right|}{c_{l}\ell_{2}\left(\tilde{\mathbf{F}}^{(2)}\right)}\\
	& = & \frac{\left|\sum_{j=1}^{O}\mathbf{W}_{l}^{(j)}\mathbf{X}_{j}^{(1)}\right|}{c_{l}}\left[\frac{k+\left|\sqrt{2\delta-\delta^{2}}\right|}{\ell_{2}\left(\tilde{\mathbf{F}}^{(2)}\right)}-\frac{1}{\ell_{2}\left(\tilde{\mathbf{F}}^{(1)}\right)}\right].
\end{eqnarray*}
Notice that $\frac{\left|\sum_{j=1}^{O}\mathbf{W}_{l}^{(j)}\mathbf{X}_{j}^{(1)}\right|}{c_{l}}<1$
since $c_{l}=\sqrt{\sum_{i}^{N}\left(\mathbf{F}_{l}^{(i)}\right)^{2}}$.
Thus:
\[
\bar{\mathbf{Z}}_{l}\leq\left[\frac{k+\left|\sqrt{2\delta-\delta^{2}}\right|}{\ell_{2}\left(\tilde{\mathbf{F}}^{(2)}\right)}-\frac{1}{\ell_{2}\left(\tilde{\mathbf{F}}^{(1)}\right)}\right].
\]
The overall Euclidean distance between the representations $\mathbf{Z}^{(1)}$
and $\mathbf{Z}^{(2)}$ can then be bounded by:
\begin{eqnarray*}
	D_{E}\left[\mathbf{Z}^{(1)},\mathbf{Z}^{(2)}\right] & = & \sqrt{\sum_{l=1}^{L}\left(\bar{\mathbf{Z}}_{l}\right)^{2}}\\
	& \leq & L\cdot\left(\frac{k+\left|\sqrt{2\delta-\delta^{2}}\right|}{\ell_{2}\left(\tilde{\mathbf{F}}^{(2)}\right)}-\frac{1}{\ell_{2}\left(\tilde{\mathbf{F}}^{(1)}\right)}\right).
\end{eqnarray*}
Thus $\epsilon=L\cdot\left(\frac{k+\left|\sqrt{2\delta-\delta^{2}}\right|}{\ell_{2}\left(\tilde{\mathbf{F}}^{(2)}\right)}-\frac{1}{\ell_{2}\left(\tilde{\mathbf{F}}^{(1)}\right)}\right)$.
$\blacksquare$

\subsection{Proposition \ref{basis-pursuit-proposition}
\label{app:basis-pursuit-proposition}}

\textbf{Proposition 2.}
\emph{	Let $\mathsf{Z}=\left\{ \mathbf{Z}^{(i)}\right\} _{i=1}^{N}$
	be a set of vectors such that $\mathbf{Z}^{(i)}\in\mathbb{R}^{L}$
	and $\sum_{j=1}^{L}\left(\mathbf{Z}_{j}^{(i)}\right)^{2}=1$. Then
	an optimal set of vectors that solve the optimization problem $\displaystyle \min_{\mathbf{Z} \in \mathbb{R}^{L \times N}} $ $\sum_{i=1}^{N}\sum_{j=1}^{L}\mathbf{Z}_{j}^{(i)}$
	is given by a multi-set of the orthonormal bases of $\mathbb{R}^{L}$.}

\textbf{Proof}. We will prove this proposition geometrically.

Let us consider the optimization problem:
\[
\min_{\mathbf{Z} \in \mathbb{R}^{L}} \sum_{j=1}^{L}\mathbf{Z}_{j}^{(1)},
\]
subject the constraint:
\[
\sum_{j=1}^{L}\left(\mathbf{Z}_{j}^{(1)}\right)^{2}=1.
\]
The constraint defines the set of points describing a unitary hyper-sphere
in $\mathbb{R}^{L}$, while the minimization problem defines diamond-shaped
level sets \citep{Bishop2006}. The minimal level set intersecting
the unitary hyper-sphere is the diamond inscribed in the unit sphere.
The intersection points constitute the solution of our minimization
problem. These points are the intersection points between the unit
hyper-sphere and the axes of $\mathbb{R}^{L}$, having a single component set to one, while all the others are set
to zero. By definition, these $1$-sparse vectors are the
orthonormal bases $\left\{ \mathbf{e}_{i}\right\} _{i=1}^{L}$. $\blacksquare$

\subsection{Proposition \ref{non-preservation-sigm} \label{app:non-preservation-sigm}}

\textbf{Proposition 3.}
\emph{
	Let us consider the sparse filtering algorithm implemented using a sigmoid non-linearity $\sigma(x)=\frac{1}{1+e^{-x}}$. 
	Let $\left\{ \mathbf{X}^{(i)}\in\mathbb{R}^{O}\right\} _{i=1}^{N}$ be a set of points in the original space $\mathbb{R}^{O}$. 
	Then, the transformations A1, A2*, A3 and A4, where A2* is the sigmoid non-linearity, do not preserve the structure of the data described neither by the Euclidean metric nor by the cosing metric.}

\textbf{Proof.} We divide this proposition in two parts and we prove each one by counterexample.

Let us focus first on the non-preservation of the Euclidean metric. 
Let us consider the case in which $\mathbf{X}^{(1)}$ is a vector such that $\mathbf{X}^{(1)}_j = 1$, $\forall j$,  $1 \leq j \leq O$, $\mathbf{X}^{(2)}$ is another vector such that $\mathbf{X}^{(2)}=2$, $\forall j$,  $1 \leq j \leq O$, $L = O$, and $\mathbf{W}=\mathbf{I}$, where $\mathbf{I}$ is the identity matrix.

The Euclidean distance between the vectors $\mathbf{X}^{(1)}$ and $\mathbf{X}^{(2)}$ is:
\[
D_{E}\left(\mathbf{X}^{(1)},\mathbf{X}^{(2)}\right)=\sqrt{\sum_{j=1}^{O}\left(1-2\right)^{2}}=\sqrt{L}.
\]

Let us now apply the transformation $f_{A1:A4}$ to the vectors $\mathbf{X}^{(1)}$ and $\mathbf{X}^{(2)}$:
\[
\begin{array}{ccc}
f_{A1}\left(\mathbf{X}^{(1)}\right)=\mathbf{I}\mathbf{X}^{(1)}=\mathbf{X}^{(1)} &  & f_{A1}\left(\mathbf{X}^{(2)}\right)=\mathbf{I}\mathbf{X}^{(2)}=\mathbf{X}^{(2)}\\

f_{A2^*}\left(\mathbf{X}^{(1)}\right)=\sigma\left(\mathbf{X}^{(1)}\right)=\mathbf{\Sigma}^{(1)} &  & f_{A2^*}\left(\mathbf{X}^{(2)}\right)=\sigma\left(\mathbf{X}^{(2)}\right)=\mathbf{\Sigma}^{(2)}\\

f_{A3}\left(\mathbf{\Sigma}^{(1)}\right)=\left[\frac{\mathbf{\Sigma}^{(1)}_j}{\sqrt{\sum_{i=1}^N \mathbf{\Sigma}^{(i)}_j }}\right] &  & f_{A3}\left(\mathbf{\Sigma}^{(2)}\right)=\left[\frac{\mathbf{\Sigma}^{(2)}_j}{\sqrt{\sum_{i=1}^N \mathbf{\Sigma}^{(i)}_j }}\right]\\

f_{A4}\left(\left[\frac{\mathbf{\Sigma}^{(1)}_j}{\sqrt{\sum_{j=1}^L \mathbf{\Sigma}^{(i)}_j }}\right] \right)=\left[\frac{1}{\sqrt{L}}\right]=\mathbf{Z}^{(1)} &  & f_{A4}\left(\left[\frac{\mathbf{\Sigma}^{(2)}_j}{\sqrt{\sum_{j=1}^L \mathbf{\Sigma}^{(i)}_j }}\right] \right)=\left[\frac{1}{\sqrt{L}}\right]=\mathbf{Z}^{(1)}.
\end{array}
\]
Thus, $f_{A1:A4}\left(\mathbf{X}^{(1)}\right)=\mathbf{Z}^{(1)}$ and $f_{A1:A4}\left(\mathbf{X}^{(2)}\right)=\mathbf{Z}^{(1)}$.
Now, the Euclidean distance between the vectors $f_{A1:A4}\left(\mathbf{X}^{(1)}\right)$
and $f_{A1:A4}\left(\mathbf{X}^{(2)}\right)$ is:
\[
D_{E}\left(\mathbf{Z}^{(1)},\mathbf{Z}^{(1)}\right)=0.
\]
Therefore the transformations from A1 to A4 do not preserve the structure of the data described by the Euclidean metric. This proves the first part of the proposition.\\

Let us focus now on the non-preservation of the cosine metric. Let us consider the case in which $\mathbf{X}^{(1)}$ is a vector such that $\mathbf{X}^{(1)}_j = {2^j}$, $\forall j$,  $1 \leq j \leq O$, $\mathbf{X}^{(2)}$ is another vector such that $\mathbf{X}^{(2)}=-\mathbf{X}^{(1)}$, $L = O = 2$, and $\mathbf{W}=\mathbf{I}$, where $\mathbf{I}$ is the identity matrix.

The cosine distance between the vectors $\mathbf{X}^{(1)}$ and $\mathbf{X}^{(2)}$ is:
\[
D_{C}\left(\mathbf{X}^{(1)},\mathbf{X}^{(2)}\right)=1-\left|\frac{\sum_{j=1}^{O}\mathbf{X}_{j}^{(1)}\mathbf{X}_{j}^{(2)}}{\sqrt{\sum_{j=1}^{O}\left(\mathbf{X}_{j}^{(1)}\right)^{2}}\sqrt{\sum_{j=1}^{O}\left(\mathbf{X}_{j}^{(2)}\right)^{2}}}\right| = 0.
\]
Let us now apply the transformation $f_{A1:A4}$ to the vectors $\mathbf{X}^{(1)}$ and $\mathbf{X}^{(2)}$:
\[
\begin{array}{ccc}
f_{A1}\left(\mathbf{X}^{(1)}\right)=\mathbf{I}\mathbf{X}^{(1)}=\mathbf{X}^{(1)} &  & f_{A1}\left(\mathbf{X}^{(2)}\right)=\mathbf{I}\mathbf{X}^{(2)}=\mathbf{X}^{(2)}\\

f_{A2^*}\left(\mathbf{X}^{(1)}\right)=\sigma\left(\mathbf{X}^{(1)}\right)=\mathbf{\Sigma}^{(1)} &  & f_{A2^*}\left(\mathbf{X}^{(2)}\right)=\sigma\left(\mathbf{X}^{(2)}\right)=\mathbf{\Sigma}^{(2)}\\

f_{A3}\left(\mathbf{\Sigma}^{(1)}\right)=\left[\frac{\mathbf{\Sigma}^{(1)}_j}{\sqrt{\sum_{i=1}^N \mathbf{\Sigma}^{(i)}_j }}\right]  &  & f_{A3}\left(\mathbf{\Sigma}^{(2)}\right)=\left[\frac{\mathbf{\Sigma}^{(2)}_j}{\sqrt{\sum_{i=1}^N \mathbf{\Sigma}^{(i)}_j }}\right]\\

f_{A4}\left(\left[\frac{\mathbf{\Sigma}^{(1)}_j}{\sqrt{\sum_{j=1}^L \mathbf{\Sigma}^{(i)}_j }}\right]\right) = \mathbf{Z}^{(1)} &  & f_{A4}\left(\left[\frac{\mathbf{\Sigma}^{(2)}_j}{\sqrt{\sum_{j=1}^L \mathbf{\Sigma}^{(i)}_j }}\right]\right)=\mathbf{Z}^{(2)}.
\end{array}
\]
Thus, $f_{A1:A4}\left(\mathbf{X}^{(1)}\right)=\mathbf{Z}^{(1)}$ and $f_{A1:A4}\left(\mathbf{X}^{(2)}\right)=\mathbf{Z}^{(2)}$, where $\mathbf{Z}^{(1)}=\left[\begin{array}{cc}
0.99 & 0.14\end{array}\right]$ and $\mathbf{Z}^{(2)}=\left[\begin{array}{cc}
0.70 & 0.71\end{array}\right]$. 
Now, the cosine distance between the vectors $f_{A1:A4}\left(\mathbf{X}^{(1)}\right)$
and $f_{A1:A4}\left(\mathbf{X}^{(2)}\right)$ is:
\[
D_{C}\left(\mathbf{Z}^{(1)},\mathbf{Z}^{(2)}\right) \neq 0.
\]
Therefore the transformations from A1 to A4 do not preserve the structure of the data described by the cosine metric. This proves the second part of the proposition. $\blacksquare$

\subsection{Proposition \ref{non-preservation-relu} \label{app:non-preservation-relu}}

\textbf{Proposition 4.}
\emph{	
	Let us consider the sparse filtering algorithm implemented using a soft ReLU non-linearity $ReLU(x)=\max\left(\epsilon,x\right)$, where $\epsilon$ is 	a small negligible value (for instance, $\epsilon=10^{-8}$). 
	Let $\left\{ \mathbf{X}^{(i)}\in\mathbb{R}^{O}\right\} _{i=1}^{N}$ be a set of points in the original space $\mathbb{R}^{O}$. 
	Then, the transformations A1, A2*, A3 and A4, where A2* is the ReLU non-linearity, do not preserve the structure of the data described neither by the Euclidean metric nor by the cosing metric.}

\textbf{Proof.} We divide this proposition in two parts and we prove each one by counterexample.

Let us focus first on the non-preservation of the Euclidean metric. Let us consider the case in which $\mathbf{X}^{(1)}$ is a vector such that $\mathbf{X}^{(1)}_j = -\frac{3}{\sqrt{2}}$, $\forall j$,  $1 \leq j \leq O$, $\mathbf{X}^{(2)}$ is another vector such that $\mathbf{X}^{(2)}=-\frac{1}{\sqrt{2}}$, $\forall j$,  $1 \leq j \leq O$, $L = O$, and $\mathbf{W}=\mathbf{I}$, where $\mathbf{I}$ is the identity matrix.

The Euclidean distance between the vectors $\mathbf{X}^{(1)}$ and $\mathbf{X}^{(2)}$ is:
\[
D_{E}\left(\mathbf{X}^{(1)},\mathbf{X}^{(2)}\right)=\sqrt{\sum_{j=1}^{O}\left(-\frac{3}{\sqrt{2}}+\frac{1}{\sqrt{2}}\right)^{2}}=\sqrt{2L}.
\]
Let us now apply the transformation $f_{A1:A4}$ to the vectors $\mathbf{X}^{(1)}$ and $\mathbf{X}^{(2)}$:
\[
\begin{array}{ccc}
f_{A1}\left(\mathbf{X}^{(1)}\right)=\mathbf{I}\mathbf{X}^{(1)}=\mathbf{X}^{(1)} &  & f_{A1}\left(\mathbf{X}^{(2)}\right)=\mathbf{I}\mathbf{X}^{(2)}=\mathbf{X}^{(2)}\\

f_{A2^*}\left(\mathbf{X}^{(1)}\right)=ReLU\left(\mathbf{X}^{(1)}\right)=\left[\epsilon\right] &  & f_{A2^*}\left(\mathbf{X}^{(2)}\right)=ReLU\left(\mathbf{X}^{(2)}\right)=\left[\epsilon\right]\\

f_{A3}\left(\left[\epsilon\right]\right)=\left[\frac{\epsilon}{\sqrt{N}\epsilon}\right]=\left[\frac{1}{\sqrt{N}}\right] &  & f_{A3}\left(\left[\epsilon\right]\right)=\left[\frac{\epsilon}{\sqrt{N}\epsilon}\right]=\left[\frac{1}{\sqrt{N}}\right]\\

f_{A4}\left(\left[\frac{1}{\sqrt{N}}\right]\right)=\left[\frac{\sqrt{N}}{\sqrt{N}\sqrt{L}}\right]=\left[\frac{1}{\sqrt{L}}\right]=\mathbf{Z}^{(1)} &  & f_{A4}\left(\left[\frac{1}{\sqrt{N}}\right]\right)=\left[\frac{\sqrt{N}}{\sqrt{N}\sqrt{L}}\right]=\left[\frac{1}{\sqrt{L}}\right]=\mathbf{Z}^{(1)}.
\end{array}
\]
Thus, $f_{A1:A4}\left(\mathbf{X}^{(1)}\right)=\mathbf{Z}^{(1)}$ and $f_{A1:A4}\left(\mathbf{X}^{(2)}\right)=\mathbf{Z}^{(1)}$.
Now, the Euclidean distance between the vectors $f_{A1:A4}\left(\mathbf{X}^{(1)}\right)$
and $f_{A1:A4}\left(\mathbf{X}^{(2)}\right)$ is:
\[
D_{E}\left(\mathbf{Z}^{(1)},\mathbf{Z}^{(1)}\right)=0.
\]
Therefore the transformations from A1 to A4 do not preserve the structure of the data described by the Euclidean metric. This proves the first part of the proposition.\\

Let us focus now on the non-preservation of the cosine metric. Let us consider the case in which $\mathbf{X}^{(1)}$ is a vector such that $\mathbf{X}^{(1)}_j = \frac{1}{2^j}$, $\forall j$,  $1 \leq j \leq O$, $\mathbf{X}^{(2)}$ is another vector such that $\mathbf{X}^{(2)}=-\mathbf{X}^{(1)}$, $L = O = 2$, and $\mathbf{W}=\mathbf{I}$, where $\mathbf{I}$ is the identity matrix.

The cosine distance between the vectors $\mathbf{X}^{(1)}$ and $\mathbf{X}^{(2)}$ is:
\[
D_{C}\left(\mathbf{X}^{(1)},\mathbf{X}^{(2)}\right)=1-\left|\frac{\sum_{j=1}^{O}\mathbf{X}_{j}^{(1)}\mathbf{X}_{j}^{(2)}}{\sqrt{\sum_{j=1}^{O}\left(\mathbf{X}_{j}^{(1)}\right)^{2}}\sqrt{\sum_{j=1}^{O}\left(\mathbf{X}_{j}^{(2)}\right)^{2}}}\right| = 0.
\]
Let us now apply the transformation $f_{A1:A4}$ to the vectors $\mathbf{X}^{(1)}$ and $\mathbf{X}^{(2)}$:
\[
\begin{array}{ccc}
f_{A1}\left(\mathbf{X}^{(1)}\right)=\mathbf{I}\mathbf{X}^{(1)}=\mathbf{X}^{(1)} &  & f_{A1}\left(\mathbf{X}^{(2)}\right)=\mathbf{I}\mathbf{X}^{(2)}=\mathbf{X}^{(2)}\\

f_{A2^*}\left(\mathbf{X}^{(1)}\right)=ReLU\left(\mathbf{X}^{(1)}\right)=\mathbf{X}^{(1)} &  & f_{A2^*}\left(\mathbf{X}^{(2)}\right)=ReLU\left(\mathbf{X}^{(2)}\right)=\left[\epsilon\right]\\

f_{A3}\left(\mathbf{X}^{(1)}\right)=\left[\frac{\mathbf{X}_{j}^{(1)}2j}{\sqrt{1+2^{2j}\epsilon^{2}}}\right]=\left[\frac{1}{\sqrt{1+2^{2j}\epsilon^{2}}}\right] &  & f_{A3}\left(\left[\epsilon\right]\right)=\left[\frac{\epsilon2j}{\sqrt{1+2^{2j}\epsilon^{2}}}\right]\\

f_{A4}\left(\left[\frac{1}{\sqrt{1+2^{2j}\epsilon^{2}}}\right]\right)=\left[\frac{\frac{1}{\sqrt{1+2^{2j}\epsilon^{2}}}}{\sqrt{\sum_{j=1}^{L}\frac{1}{1+2^{2j}\epsilon^{2}}}}\right]=\mathbf{Z}^{(1)} &  & f_{A4}\left(\left[\frac{\epsilon2j}{\sqrt{1+2^{2j}\epsilon^{2}}}\right]\right)=\left[\frac{\frac{\epsilon2j}{\sqrt{1+2^{2j}\epsilon^{2}}}}{\sqrt{\sum_{j=1}^{L}\frac{2^{2j}\epsilon^{2}}{1+2^{2j}\epsilon^{2}}}}\right]=\mathbf{Z}^{(2)}.
\end{array}
\]
Thus, $f_{A1:A4}\left(\mathbf{X}^{(1)}\right)=\mathbf{Z}^{(1)}$ and $f_{A1:A4}\left(\mathbf{X}^{(2)}\right)=\mathbf{Z}^{(2)}$, where $\mathbf{Z}^{(1)}=\left[\begin{array}{cc}
\frac{\sqrt{1+16\epsilon^{2}}}{\sqrt{2+20\epsilon^{2}}} & \frac{\sqrt{1+4\epsilon^{2}}}{\sqrt{2+20\epsilon^{2}}}\end{array}\right]$ and $\mathbf{Z}^{(2)}=\left[\begin{array}{cc}
\frac{\sqrt{1+16\epsilon^{2}}}{\sqrt{5+32\epsilon^{2}}} & \frac{2\sqrt{1+4\epsilon^{2}}}{\sqrt{5+32\epsilon^{2}}}\end{array}\right]$. 
Now, the cosine distance between the vectors $f_{A1:A4}\left(\mathbf{X}^{(1)}\right)$
and $f_{A1:A4}\left(\mathbf{X}^{(2)}\right)$ is:
\[
D_{C}\left(\mathbf{Z}^{(1)},\mathbf{Z}^{(2)}\right) \neq 0.
\]
Therefore the transformations from A1 to A4 do not preserve the structure of the data described by the cosine metric. This proves the second part of the proposition. $\blacksquare$

\subsection{Theorem \ref{bounds-for-preserving} \label{app:bounds-for-preserving}}

\textbf{Theorem 5.} 
	\emph{Let $\mathbf{X}^{(1)}\in\mathbb{R}^{O}$ be a point
	in the original space $\mathbb{R}^{O}$ and let $R_{\mathbf{X}^{(1)}}^{\mathbf{e}_{k}}$
	be a representation filter centered on $\mathbf{X}^{(1)}$, that is,
	$R_{\mathbf{X}^{(1)}}^{\mathbf{e}_{k}}\left(\mathbf{X}^{(1)}\right)=0$.
	Let us now consider a point $\mathbf{X}^{(2)}\in\mathbb{R}^{O}$ within
	the same representation cone, that is, a point such that $R_{\mathbf{X}^{(1)}}^{\mathbf{e}_{k}}\left(\mathbf{X}^{(2)}\right)\leq\epsilon$
	for an arbitrarily small $\epsilon\in\mathbb{R}$, $\epsilon>0$.}
	
	\emph{Let us assume that: (i) points $\mathbf{X}^{(i)}$ distribute in a
	limited region of space bounded by $M$; and, (ii) points $\mathbf{X}^{(i)}$
	distribute uniformly in this limited region of space.}
	
	\emph{Then, given that $R_{\mathbf{X}^{(1)}}^{\mathbf{e}_{k}}\left(\mathbf{X}^{(2)}\right)\leq\epsilon$, it follows:
	\[
	\frac{O\delta}{\left(\frac{M}{m}\right)^{O-1}}\cdot\frac{\Gamma\left(\frac{O+1}{2}\right)}{\Gamma\left(\frac{O+2}{2}\right)}\leq
	P\left(D_{E}\left[\mathbf{Z}^{(1)},\mathbf{Z}^{(2)}\right]\leq\delta\right)
	\leq\frac{O\delta}{m}\cdot\frac{\Gamma\left(\frac{O+1}{2}\right)}{\Gamma\left(\frac{O+2}{2}\right)},
	\]
	where $\delta\in\mathbb{R}$, $\delta>0$ defines the neighborhood
	of $\mathbf{X}^{(1)}$, $m$ is the distance of $\mathbf{X}^{(1)}$
	from the origin, and $\Gamma(\cdot)$ is the gamma function.}

\textbf{Proof.} Let us consider $\mathbf{X}^{(1)}\in\mathbb{R}^{O}$
and let us define its neighborhood as the set of points $\mathbf{X}^{(i)}$
within a hyper-sphere of radius $\delta$, that is, $D_{E}\left[\mathbf{X}^{(1)},\mathbf{X}^{(i)}\right]\leq\delta$.

\begin{figure}
	\begin{centering}
		\includegraphics[scale=0.9]{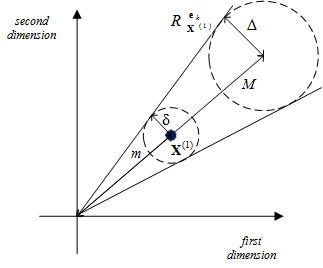} 
		\par\end{centering}
	
	\caption{Schema of the data point $\mathbf{X}^{(1)}$, the neighborhood
		of $\mathbf{X}^{(1)}$, and the representation filter $R_{\mathbf{X}^{(1)}}^{\mathbf{e}_{k}}$
		in two-dimensional space. \label{fig:Hypercone}}
\end{figure}

Let us consider now the representation filter $R_{\mathbf{X}^{(1)}}^{\mathbf{e}_{k}}$
and let $m$ be the distance of $\mathbf{X}^{(1)}$ from the origin.
We first define the minimal representation filter $R_{\mathbf{X}^{(1)}}^{\mathbf{e}_{k}}$
as the hyper-cone of height $m$ and radius $\delta$ inscribing the
neighborhood of $\mathbf{X}^{(1)}$. We also define a maximal representation
filter $R_{\mathbf{X}^{(1)}}^{\mathbf{e}_{k}}$ as the hyper-cone
of height $M$ and, by trigonometry, radius $\Delta=M\cdot\frac{\delta}{m}$.
For illustration, refer to the schema in Figure \ref{fig:Hypercone},
where we represented this setup in the case $O=2$.

Let us now consider the point $\mathbf{X}^{(2)}$ sampled within the
representation filter $R_{\mathbf{X}^{(1)}}^{\mathbf{e}_{k}}$. Since
the sampling probability is uniform within the representation filter
$R_{\mathbf{X}^{(1)}}^{\mathbf{e}_{k}}$, we can evaluate the probability
of $\mathbf{X}^{(2)}$ to fall in the neighborhood of $\mathbf{X}^{(1)}$
as the volume of the neighborhood of $\mathbf{X}^{(1)}$ normalized
by the total volume of the representation filter $R_{\mathbf{X}^{(1)}}^{\mathbf{e}_{k}}$. 

Let us consider the neighborhood of $\mathbf{X}^{(1)}$. Its volume
can be computed as: 
\[
V_{sphere}(O,\delta)=\mathcal{V}_{O}\delta^{O},
\]
where $\mathcal{V}_{O}$ is the following function:
\[
\mathcal{V}_{n}=\frac{\pi^{\frac{n}{2}}}{\Gamma\left(\frac{n}{2}+1\right)}.
\]
Let us now consider the representation filter $R_{\mathbf{X}^{(1)}}^{\mathbf{e}_{k}}$.
We bound this volume considering the minimal and maximal hyper-cone
described above. The volume of the hyper-cone depends on the volume
of the lower-dimensional hyper-sphere in the base \citep{Ball1997}
and it can be computed as:
\begin{eqnarray*}
	\frac{1}{O}\cdot m\cdot V_{sphere}(O-1,\delta) & \leq V_{cone}(O,\delta,l)\leq & \frac{1}{O}\cdot M\cdot V_{sphere}(O-1,\Delta)\\
	\frac{1}{O}\cdot m\cdot\mathcal{V}_{O-1}\cdot\delta^{O-1} & \leq V_{cone}(O,\delta,l)\leq & \frac{1}{O}\cdot M\cdot\mathcal{V}_{O-1}\cdot\left(M\cdot\frac{\delta}{m}\right)^{O-1}
\end{eqnarray*}
Let us now consider the ratio of the volume of the hyper-sphere and
the volume of the hyper-cone:
\begin{eqnarray*}
	\frac{\mathcal{V}_{O}\delta^{O}}{\frac{1}{O}\cdot M\cdot\mathcal{V}_{O-1}\cdot\left(M\cdot\frac{\delta}{m}\right)^{O-1}} & \leq\frac{V_{sphere}(O,\delta)}{V_{cone}(O,\delta,l)}\leq & \frac{\mathcal{V}_{O}\delta^{O}}{\frac{1}{O}\cdot m\cdot\mathcal{V}_{O-1}\cdot\delta^{O-1}}\\
	\frac{O\delta m^{O-1}}{M^{O}}\cdot\frac{\Gamma\left(\frac{O+1}{2}\right)}{\Gamma\left(\frac{O+2}{2}\right)} & \leq\frac{V_{sphere}(O,\delta)}{V_{cone}(O,\delta,l)}\leq & \frac{O\delta}{m}\cdot\frac{\Gamma\left(\frac{O+1}{2}\right)}{\Gamma\left(\frac{O+2}{2}\right)}
\end{eqnarray*}

Thus, since we assumed that $R_{\mathbf{X}^{(1)}}^{\mathbf{e}_{k}}\left(\mathbf{X}^{(2)}\right)\leq\epsilon$, it follows that $\frac{O\delta}{\left(\frac{M}{m}\right)^{O-1}}\cdot\frac{\Gamma\left(\frac{O+1}{2}\right)}{\Gamma\left(\frac{O+2}{2}\right)}\leq
P\left(D_{E}\left[\mathbf{Z}^{(1)},\mathbf{Z}^{(2)}\right]\leq\delta\right)
\leq\frac{O\delta}{m}\cdot\frac{\Gamma\left(\frac{O+1}{2}\right)}{\Gamma\left(\frac{O+2}{2}\right)}$.
$\blacksquare$

\section*{Acknowledgments}
The authors would like to thank the anonymous reviewers for their insightful comments that improved the presentation of this manuscript. The authors are also grateful to William Woof for personal communication. 
F. M. Zennaro's work was supported by the Kilburn PhD studentship.

\section*{References}
\bibliography{zennaro16a}

\end{document}